\documentclass{article}
\usepackage{iclr2026_conference,times}

\usepackage{arydshln}
\usepackage[utf8]{inputenc}
\usepackage[T1]{fontenc}
\usepackage{hyperref}
\usepackage{url}
\usepackage{booktabs}
\usepackage{amsfonts}
\usepackage{nicefrac}
\usepackage{microtype}
\usepackage[dvipsnames]{xcolor}
\usepackage{graphicx}
\usepackage{multirow}
\usepackage{multicol}
\usepackage{mathtools}
\usepackage{amsmath}
\usepackage{tabularx}
\usepackage{array}
\usepackage{algorithm}
\usepackage{algpseudocode}
\usepackage{amssymb}
\usepackage{colortbl}
\usepackage{tikz}
\usepackage{wrapfig}
\usepackage{xspace}
\usepackage[most]{tcolorbox}
\tcbuselibrary{breakable}

\usetikzlibrary{
    positioning,
    matrix,
    fit,
    calc,
    decorations.pathreplacing,
    backgrounds,
    shapes.misc
}

\usepackage{amsmath,amsfonts,bm}

\def\eqref#1{equation~\ref{#1}}

\def\1{\bm{1}}

\DeclareMathAlphabet{\mathsfit}{\encodingdefault}{\sfdefault}{m}{sl}
\SetMathAlphabet{\mathsfit}{bold}{\encodingdefault}{\sfdefault}{bx}{n}

\newcommand{\methodl}{\textsc{Reinforcement Learning with Comparative Evidence}\xspace}

\newcommand{\method}{\textsc{RLCE}\xspace}

\newcolumntype{Y}{>{\centering\arraybackslash}X}

\definecolor{gainbg}{HTML}{CFE6D4}
\definecolor{flatbg}{HTML}{F6EFCD}
\definecolor{lossbg}{HTML}{F2D6D3}

\definecolor{boxblue}{RGB}{91,122,153}
\definecolor{boxblueback}{RGB}{247,249,251}

\newtcolorbox{promptbox}[2][]{%
    enhanced,
    breakable,
    title=#2,
    colback=boxblueback,
    colframe=boxblue,
    colbacktitle=boxblue,
    coltitle=white,
    fonttitle=\bfseries,
    halign title=flush left,
    fontupper=\small\ttfamily\raggedright,
    arc=2.5mm,
    boxrule=0.9pt,
    left=3mm,
    right=3mm,
    top=2.5mm,
    bottom=2.5mm,
    toptitle=1mm,
    bottomtitle=1mm,
    parbox=false,
    before upper={
        \setlength{\parskip}{4pt}
        \setlength{\parindent}{0pt}
    },
    #1
}

\newcommand{\dpos}[2]{\cellcolor{green!#2}$+#1$}
\newcommand{\dneg}[2]{\cellcolor{red!#2}$-#1$}

\usepackage[most]{tcolorbox}
\usepackage{enumitem}

\title{Reinforcement Learning with Comparative \\ Evidence for Social Intelligence}
\author{
\hspace{-2.6pt}
Keane Ong\textsuperscript{1,2}\thanks{Equal contribution} \ ,
\ Yuriel Ryan\textsuperscript{2}\footnotemark[1] \ ,
\ Sabri Boughorbel\textsuperscript{3},
\ Vladimir Necula\textsuperscript{4}, \\
\bfseries
Jack Wei Lun Shi\textsuperscript{2},
\ Rui Mao\textsuperscript{5},
\ Roy Ka-Wei Lee\textsuperscript{6},
\ Adriel Kuek\textsuperscript{7}, \\
\bfseries
Nancy F. Chen\textsuperscript{8},
\ Erik Cambria\textsuperscript{5},
\ Gianmarco Mengaldo\textsuperscript{2},
\ Paul Pu Liang\textsuperscript{1} \\
\textsuperscript{1}MIT \quad
\textsuperscript{2}NUS \quad
\textsuperscript{3}Prince Sattam bin Abdulaziz University \quad
\textsuperscript{4}University of Cambridge \\
\textsuperscript{5}NTU \quad
\textsuperscript{6}UBC \quad
\textsuperscript{7}DSO National Laboratories \quad
\textsuperscript{8}A*STAR
}
\iclrfinalcopy

\newif\ifwithappendix

\withappendixtrue

\ifwithappendix
\else
\fi

\begin{document}

\maketitle


\begin{abstract}
Developing socially intelligent AI remains heavily dependent on human-annotated data, limiting the scale and breadth of social understanding that models can acquire. Methods that derive training signals from unlabeled data offer a path beyond this dependence, but social predictions lack the verification oracles available in domains such as mathematics and coding. Moreover, core social targets such as affect, intent, preference, and pragmatic meaning are often ambiguous. The same observed behavior can support multiple plausible interpretations, making it difficult to verify which is best supported. To address this challenge, we introduce \textbf{\methodl (\textsc{RLCE})}, a reinforcement learning method that learns social understanding from unlabeled training data without constructing rewards from ground-truth annotations. Given distinct answers in a rollout group, \method constructs \emph{evidence tests} that identify observable evidence favoring an answer over another, validates these tests against the input sample, and aggregates test outcomes to determine the best-supported interpretation. Evidence tests are then regenerated as the policy produces new answers during training, enabling tests to evolve with the policy's answers. Across four benchmarks spanning affect, pragmatics, communicative intent, and preference, \method attains the strongest performance among seven methods that use no ground-truth training labels for rewards, including consensus, policy LLM-judge verification, multimodal co-evolution, and rubric-based rewards. Gains over the strongest baseline are +3.58 on MOSEI, +4.50 points on UR-FUNNY, +1.15 on IntentQA, and +18.93 on PrefEval. Analyses further show that \method exhibits a larger share of reward variation between correct and incorrect predictions than compared rubric methods, can overturn erroneous policy-derived preferences, and benefits from pairwise test construction, compositional test aggregation, and on-policy test evolution. Code and models will be released after the review process.
\end{abstract}





\section{Introduction}






Building machines with social intelligence, the ability to understand people, their affective and cognitive states, preferences, and behavior, has been one of AI’s longest-standing and most fundamental ambitions~\citep{picard2000affective, breazeal2000sociable}. However, despite the remarkable progress of modern AI, socially intelligent models remains heavily reliant on human-annotated data~\citep{zhang2026affectivescarce}, which is difficult to obtain at scale because human behavior is highly heterogeneous, context-dependent, and ambiguous to interpret~\citep{zhang2026affectivescarce, ong2026omnisapiens}. This places a fundamental ceiling on social intelligent AI systems, limiting what models can learn and how well they generalize to new settings~\citep{ong2026humanbehavioratlas}. Raising this ceiling, we argue, requires \emph{self-evolving social intelligence}: systems that derive learning signals from unlabeled training data rather than human annotations. Yet, this remains challenging as social predictions lack a verification oracle (e.g., a proof checker), unlike solutions in mathematics and coding~\citep{zhao2026absolute}. Consequently, self-generated supervision risks reinforcing errors~\citep{arazo2020pseudo}. 




To mitigate this problem, prior work has explored proxy verification signals in place of a verification oracle, including LLM judgments~\citep{yuan2024self}, self-consistency~\citep{chen2023universal, he2025visplay, zhang2026video}, and rubric-based rewards~\citep{shao2025dr, gunjal2025rubrics, wu2025rlac}. However, social understanding poses a distinct verification problem: the social targets to be modeled - e.g., intent, sentiment, preference, or pragmatic meaning, are ambiguous, and manifest through observable language, expressions, cues, and behavior~\citep{mathur2024socialagents, vinciarelli2009social}. These observations can often be compatible with multiple plausible interpretations, although only one corresponds to the underlying social target~\citep{vinciarelli2009social,vinciarelli2011bridging}. Existing proxy verification signals do not explicitly resolve among these interpretations. Rubric-based methods typically evaluate each response independently against general criteria, which may leave multiple interpretations similarly plausible~\citep{gunjal2025rubrics,shao2025dr}. Additionally, consensus-based verification may favor agreement across predictions rather than what is actually observed in the sample~\citep{chen2023universal,he2025visplay,zhang2026video}.

To address this gap, we introduce \textbf{\methodl (\method)}, an RL method for learning social understanding from unlabeled training data. Our core principle is to verify ambiguous answers through comparative evidence: comparing answers pairwise through multiple evidence tests, each of which checks the sample for evidence (i.e., observable support) favoring an answer over another, and then aggregating the test outcomes into a verdict. This allows verification to distinguish among different interpretations of the social target based on the evidence present in the sample. Given answers from a rollout group, \method constructs evidence tests that check for evidence favoring an answer over another. For example, if “Nice job” could mean \emph{praise} or \emph{sarcasm}, rather than checking a generic cue such as whether the speaker smiles, which may occur in either case, \method constructs a test around a distinction that helps resolve the two interpretations, such as whether the speaker’s tone aligns with or contradicts the positive words. Each test is then evaluated against the sample's content to determine whether the specified evidence is present, and the resulting outcomes are aggregated to identify the best-supported answer, which receives the reward for GRPO. As the policy updates, \method regenerates evidence tests for its latest answers, keeping them specific to the policy's outputs.



We evaluate \method on four established multimodal and text benchmarks spanning core and diverse social understanding tasks: affective understanding~\citep{zadeh2018multimodal}, pragmatic reasoning~\citep{hasan2019ur}, communicative-intent~\citep{li2023intentqa}, and preference and personalization~\citep{zhao2025llms}; comparing against state-of-the-art methods that construct rewards from unlabeled training data, including consensus-based approaches, from self-consistency~\citep{chen2023universal} to multimodal evolution~\citep{he2025visplay,zhang2026video}, policy LLM-Judge~\citep{yuan2024self}, and rubric-based rewards~\citep{gunjal2025rubrics, shao2025dr,wu2025rlac}. Our experiments show: (1) relative to baselines, \method achieves the best mean performance across benchmarks, with gains over the strongest baseline of +1.15 (IntentQA), +3.58 (MOSEI), +4.50 (UR-FUNNY), and +18.93 (PrefEval) points; (2) relative to compared rubric-based methods, \method exhibits a larger share of reward variation between correct and incorrect predictions, and more often overturns erroneous consensus and self-judgment preferences; (3) ablations of \method's components show that pairwise evidence-test construction improves performance, aggregating multiple tests improves verdict accuracy, and evolving tests outperform fixed tests, particularly for free-form policy outputs.

\begin{tcolorbox}[
    colback=green!6,
    colframe=green!6,
    boxrule=0pt,
    arc=3pt,
    left=8pt,
    right=8pt,
    top=6pt,
    bottom=6pt
]
\textbf{Main Contributions:}
\begin{enumerate}[leftmargin=*, itemsep=1pt, topsep=2pt, parsep=0pt]

    \item \textbf{\method for learning social understanding from unlabeled data.}
    We introduce \methodl (\method), an RL method that constructs rewards without human-provided training labels. It verifies answers through pairwise evidence tests that check the input sample for observable evidence favoring an answer over another, and aggregates their outcomes to identify the best-supported answer.

    \item \textbf{Consistent gains across four social understanding benchmarks.}
    Across four diverse social understanding tasks, \method achieves the best performance among seven label-free reward methods, exceeding the strongest baseline by up to +18.93 points.

    \item \textbf{Analysis of \method's learning signal and design.}
    \method better distinguishes correct from incorrect predictions and more often overturns erroneous consensus and self-judgment preferences. Ablations show benefits from pairwise tests, compositional aggregation, and on-policy test evolution.

\end{enumerate}
\end{tcolorbox}

\section{Related Work}

\noindent\textbf{Self-generated verification signals.}
A central challenge in learning from unlabeled training data is constructing reliable learning signals without human-provided labels. Early approaches derive this signal from agreement across the model's sampled predictions. Self-Consistency~\citep{wang2022selfconsistency} and Universal Self-Consistency~\citep{chen2023universal} use agreement across sampled answers, an approach extended to multimodal self-training by VisPlay~\citep{he2025visplay} and Video-Zero~\citep{zhang2026video}. TTRL~\citep{zuo2026ttrl} uses majority-vote pseudo-labels for policy updates, while SRLM~\citep{yuan2024self} relies on direct policy self-judgment. Recent work instead constructs explicit criteria for evaluating responses. Rubrics as Rewards (RaR)~\citep{gunjal2025rubrics} uses fixed criteria, while subsequent methods construct or refine criteria adaptively from model-generated responses~\citep{shao2025dr,guan2026evorubric,wang2026serpo,liu2026deltarubric,wu2025rlac}. By contrast, \method constructs pairwise evidence tests for candidate answers, checks whether the corresponding evidence is present in the input sample, and aggregates their outcomes across answers to identify the best-supported answer. Existing verification approaches have been explored less extensively for social understanding, where the underlying target is ambiguous and the same observed behavior can admit multiple plausible interpretations. This motivates \method's focus on resolving ambiguity by testing which interpretation is more strongly supported by the observed input.

\noindent\textbf{Socially intelligent AI.}
Core to social intelligence is inferring latent affective, intentional, preferential, and pragmatic states from observable audiovisual and linguistic behavior. Existing work has largely relied on supervised annotations, ranging from sentiment and emotion labels in MOSEI~\citep{zadeh2018multimodal} to human-authored reasoning traces in Social Genome~\citep{mathur2025socialgenome}. Recent generalist models similarly learn social understanding through supervised training on annotated data~\citep{ong2026omnisapiens,ong2026humanbehavioratlas}. Together, these approaches have advanced social understanding but retain a common dependence on human annotations for supervision. This limits how readily learning can scale to naturally occurring social data without training labels.



\begin{figure*}[h]
    \centering
    \includegraphics[
        width=\textwidth
    ]{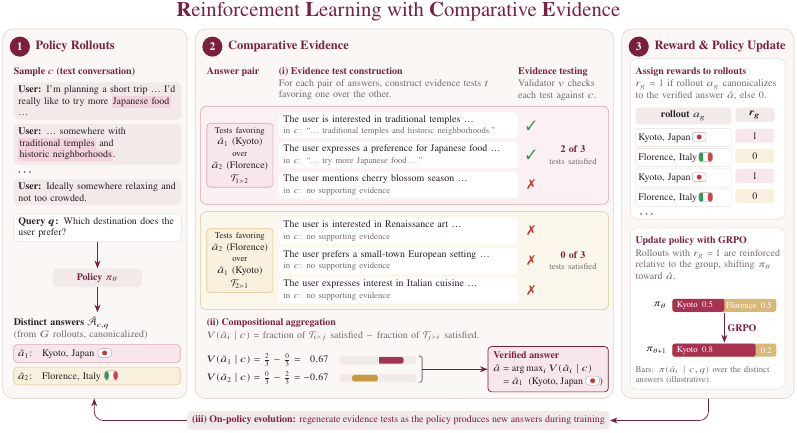}
    \vspace{-2pt}
    \caption{
    \textbf{Method Overview.} \methodl (\method) verifies candidate answers by comparing them pairwise with multiple tests that check the sample for support favoring an answer over another, and aggregating the test outcomes into a verdict. \method instantiates this through three steps: (i) \emph{Evidence-test construction:} constructing pairwise evidence tests that check for observable evidence favoring an answer over another, (ii) \emph{Compositional aggregation:} validating and aggregating these tests across answers to identify the best-supported answer for reward assignment, and (iii) \emph{On-policy evolution:} regenerating evidence tests as the policy is updated and produces new answers during training.
    }
    \label{fig:rlcv_overview}
\end{figure*}

\section{Reinforcement Learning with Comparative Evidence}
\label{sec:method}

\subsection{Preliminaries}
\label{sec:prelim}


\noindent\textbf{Problem definition: Learning social understanding from unlabeled training data.}
We consider critic-free reinforcement learning (e.g., GRPO~\citep{shao2024deepseekmath}) for social understanding tasks, where policy training and reward construction use no human-provided ground-truth annotations from the training data. Let $\pi_\theta$ denote the policy to be learnt, instantiated as an LLM. Given a multimodal sample $c$ and question $q=(u,\mathcal{A})$, where $u$ denotes the natural-language query and $\mathcal A$ the answer space, we sample $G$ rollouts $a_g \sim \pi_\theta(\cdot \mid c,q)$, $a_g \in \mathcal{A}$, $g=1,\ldots,G$. The goal is to assign rewards using only $c$ and $q$, without access to the ground-truth answer for the training example. Here $a_g$ represents an interpretation of a social target (e.g., sentiment, intent, preference, or humor) inferred from observable language, expressions, and behavior. Because these cues can be ambiguous, the same observation may support multiple plausible interpretations.



\noindent\textbf{Method overview.} \methodl (\method) verifies candidate answers through three components. (i) \emph{Evidence-test construction}: for pairs of distinct answers in a rollout group, \method constructs evidence tests that check for observable evidence in $c$ favoring an answer over the other. (ii) \emph{Compositional aggregation}: each test is validated against $c$, and the resulting test outcomes are aggregated across answer pairs into a verification score that determines the best-supported answer. (iii) \emph{On-policy evolution}: as the policy updates and produces new answers, the evidence tests are regenerated so that verification stays specific to the policy's latest answers. Fig.~\ref{fig:rlcv_overview} provides an overview, Sec.~\ref{sec:analysis},~\ref{sec:ablations}, analyzes and ablates (i), (ii) and (iii).

\subsection{Verification with Comparative Evidence}\label{sec:verification} 

\noindent\textbf{Definition (Evidence test).}
An \emph{evidence test} \(t\) checks whether the sample \(c\) contains observable evidence favoring one candidate answer over another. For example, if an answer predicts positive sentiment and another predicts negative sentiment, a test may check whether the clip shows the speaker smiling or laughing, supporting the positive interpretation over the negative one.

\textbf{Evidence test construction.}
At training iteration $\ell$, we sample a rollout group
$\{a_g^{(\ell)}\}_{g=1}^G \sim \pi_{\theta_\ell}(\cdot\mid c,q)$ and
canonicalize the rollouts into the distinct answer set
$\hat{\mathcal A}^{(\ell)}_{c,q}= \{\tilde a_1^{(\ell)},\ldots,\tilde a_K^{(\ell)}\}$, merging semantically equivalent outputs. Throughout our method, $\pi_f$ denotes a \emph{frozen} copy of the initial policy $\pi_{\theta_0}$ which is not updated during training. For open-ended outputs, canonicalization is performed by a single greedy call to $\pi_f$. For closed-label outputs, no greedy call is required for canonicalization; each rollout is mapped directly to its answer label. \method then constructs comparative evidence tests over a distinct candidate answer set $\hat{\mathcal A}^{(\ell)}_{c,q}\subseteq\mathcal A$. Concretely, for each pair of distinct answers
$(\tilde{a}_i^{(\ell)},\tilde{a}_j^{(\ell)})$, $i<j$, in the set, \method constructs one set of evidence tests for observable evidence supporting
$\tilde{a}_i^{(\ell)}$ over $\tilde{a}_j^{(\ell)}$, and another supporting
$\tilde{a}_j^{(\ell)}$ over $\tilde{a}_i^{(\ell)}$:
\begin{equation}
T_{\tilde{a}_i^{(\ell)} \succ \tilde{a}_j^{(\ell)}}
\sim
\pi_f\!\left(
\cdot \mid
c,q,\tilde{a}_i^{(\ell)},\tilde{a}_j^{(\ell)}
\right),
\qquad
T_{\tilde{a}_j^{(\ell)} \succ \tilde{a}_i^{(\ell)}}
\sim
\pi_f\!\left(
\cdot \mid
c,q,\tilde{a}_j^{(\ell)},\tilde{a}_i^{(\ell)}
\right).
\end{equation}
For example, if two answers interpret a speaker's sentiment as \emph{positive}
versus \emph{negative}, the corresponding tests separately check for observable
evidence supporting positive over negative sentiment and negative over positive
sentiment. Each test in $T_{\tilde{a}_i^{(\ell)} \succ \tilde{a}_j^{(\ell)}}$
is an observable claim whose satisfaction supports
$\tilde{a}_i^{(\ell)}$ over $\tilde{a}_j^{(\ell)}$, i.e., it provides supporting evidence for
$\tilde{a}_i^{(\ell)}$ and contradicting evidence for
$\tilde{a}_j^{(\ell)}$. Each distinct answer is thus associated with tests that support it,
$T_{\tilde{a}_i^{(\ell)}}^{+}
=
\bigcup_{j\neq i}
T_{\tilde{a}_i^{(\ell)} \succ \tilde{a}_j^{(\ell)}}$,
and tests that contradict it,
$T_{\tilde{a}_i^{(\ell)}}^{-}
=
\bigcup_{j\neq i}
T_{\tilde{a}_j^{(\ell)} \succ \tilde{a}_i^{(\ell)}}$. 

\noindent\textbf{Evidence testing.}
Given an evidence test
$t \in T_{\tilde{a}_i^{(\ell)}}^{+} \cup T_{\tilde{a}_i^{(\ell)}}^{-}$
associated with an answer $\tilde{a}_i^{(\ell)}$, validation determines whether
the claim specified by $t$ is satisfied by the sample $c$. We use a single validator $v$ to validate whether each constructed test is supported by the input. \method is validator-agnostic; in our experiments, we use $\pi_f$ as
$v$, following prior work demonstrating the effectiveness of policy-based self-judgment~\citep{yuan2024self}, while ensuring that performance improvements can be attributed to \method's verification procedure rather than a stronger external model. Accordingly, for
each evidence test $t$, the validator returns a binary satisfaction outcome:
\begin{equation}
z(t) = v(t,c), \qquad z(t) \in \{0,1\},
\end{equation}
where $z(t)=1$ indicates that $t$ is satisfied by $c$, and $z(t)=0$ otherwise. Consistent with the general principle of abstaining rather than forcing a decision under uncertainty~\citep{tomani2024uncertainty}, the validator may abstain when the sample $c$ is insufficient to determine satisfaction or non-satisfaction of a test. Abstained tests receive no outcome and are excluded from aggregation. 

\noindent\textbf{Compositional aggregation.}
\method aggregates the satisfaction outcomes of the supporting and contradicting evidence tests associated with each answer to compute its verification score, $V(\tilde{a}_i^{(\ell)}\mid c)$. Let $n_+$ and $n_-$ denote the numbers of evidence tests supporting and contradicting $\tilde{a}_i^{(\ell)}$, respectively, which are equal by construction, barring abstentions. The verification score is:
\begin{equation}
V\!\left(\tilde{a}_i^{(\ell)} \mid c\right)
=
\frac{1}{n_+}
\sum_{t \in T_{\tilde{a}_i^{(\ell)}}^{+}} z(t)
-
\frac{1}{n_-}
\sum_{t \in T_{\tilde{a}_i^{(\ell)}}^{-}} z(t).
\label{eq:candidate_verification}
\end{equation}
The two terms are the proportions of satisfied supporting and contradicting
evidence tests, respectively; their difference measures how strongly the
validated evidence-test outcomes support $\tilde{a}_i^{(\ell)}$ relative to the other distinct answers, given the observed sample $c$. Intuitively, a larger
$V(\tilde{a}_i^{(\ell)}\mid c)$ indicates stronger net evidence in favor of
$\tilde{a}_i^{(\ell)}$ over alternative answers. We therefore select the distinct answer with the highest verification score as the verified answer:
\begin{equation}
\hat{a}^{(\ell)}
=
\arg\max_{\tilde{a}_i^{(\ell)}\in\widehat{\mathcal{A}}_{c,q}^{(\ell)}}
V\!\left(\tilde{a}_i^{(\ell)}\mid c\right).
\label{eq:verified_answer}
\end{equation}


The aggregation and selection in
Eqs.(\ref{eq:candidate_verification}-\ref{eq:verified_answer}) depend on
multiple validated evidence tests. As a
result, even if some tests are incorrect, ambiguous, or uninformative, the aggregate verification score can remain accurate provided that the remaining
tests are reliable. Additionally, verification operates on distinct answers; repeated rollouts are canonicalized before test construction, preventing answer frequency from affecting $V$.

\textbf{Ties.}
If there is no clear winner with the highest verification score $V$, \method requests up to two additional evidence tests for the unresolved comparisons. If the tie remains, or if all rollouts correspond to the same answer (and thus yield no group-relative gradient), no rewards are assigned.


\noindent\textbf{On-policy evolution.}
As the policy updates and produces new answers, \method regenerates evidence tests for the new answers rather than reusing tests from earlier iterations. The
constructor and validator $\pi_f$ remain frozen, so only the evidence tests
evolve with the policy's answers, and no gradient computation occurs through
$\pi_f$. Section~\ref{sec:ablations} shows that this is particularly beneficial
for free-form outputs, whose answer sets change more substantially during training.

\subsection{Policy Update}

\noindent\textbf{Reward assignment.}
Recall that semantically equivalent rollout
answers are canonicalized into the distinct answer set
$\widehat{\mathcal{A}}_{c,q}^{(\ell)}$, from which the verified answer
$\hat{a}^{(\ell)}$ is selected in Eq.(\ref{eq:verified_answer}). We assign a
reward of $1$ to each rollout whose answer was canonicalized into
$\hat{a}^{(\ell)}$, and $0$ otherwise:
\begin{equation}
r_g^{(\ell)}
=
\begin{cases}
1, & \text{if the canonicalized answer of } a_g^{(\ell)}
      \text{ equals } \hat{a}^{(\ell)},\\
0, & \text{otherwise.}
\end{cases}
\label{eq:reward}
\end{equation}
In practice, an auxiliary format reward is added before policy optimization;
for notational simplicity, we continue to denote the resulting reward by
$r_g^{(\ell)}$. 



\noindent\textbf{Group Relative Policy Optimization.}
We optimize the policy with GRPO~\citep{shao2024deepseekmath}. For a group of $G$ rollouts with rewards $\{r_g^{(\ell)}\}_{g=1}^{G}$, and reward mean $\mu_{\mathcal{G}}^{(\ell)}$ and standard deviation $\sigma_{\mathcal{G}}^{(\ell)}$, the group-relative advantage is $A_g^{(\ell)}=(r_g^{(\ell)}-\mu_{\mathcal{G}}^{(\ell)})/\sigma_{\mathcal{G}}^{(\ell)}$, with $A_g^{(\ell)}=0$ when $\sigma_{\mathcal{G}}^{(\ell)}=0$; advantages are used in the standard clipped GRPO objective (all rewards and hyperparameters are in App.~\ref{app:rlcv-grpo}.)

\section{Experiments}\label{sec:expts}

\begin{table}[t]
\small
\centering
\caption{\textbf{Held-out test performance after training on unlabeled training data.} All methods start from the same Qwen3.5-9B base model, and we report the mean over three training runs with different random seeds. For IntentQA, MOSEI, and UR-FUNNY, which are obtained from Human Behavior Atlas~\citep{ong2026humanbehavioratlas}, we follow its evaluation setup and report LLM-Judge accuracy, binary weighted F1 (F1w2), and weighted F1, respectively. For PrefEval, we follow its standard evaluation protocol and report LLM-Judge accuracy, macro-averaged over its three preference tasks. $\Delta$ denotes change from the base model (green/red: improvement/regression), and $\uparrow/\downarrow$ denote higher/lower is better. Overall, \method achieves the most consistent performance across the baseline methods.}
\renewcommand{\arraystretch}{1.1}
\setlength{\tabcolsep}{0.3pt}
\resizebox{\linewidth}{!}{%
\begin{tabular}{lcccccccccc}
\toprule
\textbf{Method}
& \textbf{IntentQA} ($\uparrow$) & $\Delta$
& \textbf{MOSEI} ($\uparrow$) & $\Delta$
& \textbf{UR-FUNNY} ($\uparrow$) & $\Delta$
& \textbf{PrefEval}$^{\dagger}$ ($\uparrow$) & $\Delta$
& \textbf{Avg. Rank} ($\downarrow$) \\
\midrule
\rowcolor{blue!6}
Qwen3.5-9B \emph{(base model)}
& 55.25 & --
& 64.86 & --
& 63.11 & --
& 44.72 & --
& -- \\
\midrule
Universal Self-Consistency~\citep{chen2023universal}
& 56.77 & \dpos{1.52}{40}
& 66.84 & \dpos{1.98}{40}
& 59.86 & \dneg{3.25}{55}
& 41.37 & \dneg{3.35}{55}
& 5.75 \\
Policy LLM-Judge~\citep{yuan2024self}
& \underline{57.33} & \dpos{2.08}{40}
& \underline{67.35} & \dpos{2.49}{40}
& 60.44 & \dneg{2.67}{40}
& 46.73 & \dpos{2.01}{40}
& \underline{4.00} \\
Video-Zero~\citep{zhang2026video}
& 55.06 & \dneg{0.19}{15}
& 64.80 & \dneg{0.06}{15}
& 62.91 & \dneg{0.20}{15}
& 43.30 & \dneg{1.42}{25}
& 6.75 \\
VisPlay~\citep{he2025visplay}
& 55.34 & \dpos{0.09}{15}
& 65.00 & \dpos{0.14}{15}
& \underline{64.96} & \dpos{1.85}{40}
& 42.55 & \dneg{2.17}{25}
& 5.75 \\
Rubrics as Rewards~\citep{gunjal2025rubrics}
& 55.60 & \dpos{0.35}{15}
& 65.03 & \dpos{0.17}{15}
& 64.72 & \dpos{1.61}{40}
& \underline{56.45} & \dpos{11.73}{70}
& 4.25 \\
RLER~\citep{shao2025dr}
& 56.31 & \dpos{1.06}{25}
& 65.19 & \dpos{0.33}{15}
& 64.52 & \dpos{1.41}{25}
& 56.12 & \dpos{11.40}{70}
& 4.25 \\
RLAC~\citep{wu2025rlac}
& 56.75 & \dpos{1.50}{40}
& 67.15 & \dpos{2.29}{40}
& 61.96 & \dneg{1.15}{25}
& 55.27 & \dpos{10.55}{70}
& 4.25 \\
\midrule
\rowcolor{violet!8}
\textbf{\method (ours)}
& \textbf{58.48} & \dpos{3.23}{55}
& \textbf{70.93} & \dpos{6.07}{70}
& \textbf{69.46} & \dpos{6.35}{70}
& \textbf{75.38} & \dpos{30.66}{70}
& \textbf{1.00} \\
\bottomrule
\end{tabular}
}
\label{tab:main_results}
\end{table}

\textbf{Benchmarks \& baselines.} We evaluate \method on four core dimensions of
social understanding: affect using the sentiment analysis task from MOSEI~\citep{zadeh2018multimodal}, pragmatics
with UR-FUNNY~\citep{hasan2019ur}, intent with IntentQA~\citep{li2023intentqa},
and preference with PrefEval~\citep{zhao2025llms}. We compare against seven methods that construct rewards from
unlabeled training data: rubric-based methods, including Rubrics as
Rewards (RaR)~\citep{gunjal2025rubrics}, RLER~\citep{shao2025dr}, and
RLAC~\citep{wu2025rlac}; multimodal evolution methods,
Video-Zero~\citep{zhang2026video} and VisPlay~\citep{he2025visplay};
self-agreement via Universal Self-Consistency (USC)~\citep{chen2023universal}; and
a frozen Policy LLM-Judge baseline, adapted from the policy-as-judge formulation
of self-rewarding language models~\citep{yuan2024self}.

\textbf{Experimental setup.}
For MOSEI, UR-FUNNY, and IntentQA, we follow the splits, prompts, and metrics of the Human Behavior Atlas benchmark~\citep{ong2026humanbehavioratlas}; for PrefEval, we follow the evaluation setup introduced in the original work~\citep{zhao2025llms} across its explicit, choice, and persona tasks. All methods use the same Qwen3.5-9B initialization, with no ground-truth training labels used for reward construction. From Table~\ref{tab:main_results}, we report the mean over three training runs with different random seeds. Full details of all baseline implementations and the experimental setup are provided in App.~\ref{apx:baselines}.




\subsection{Results}








\textbf{Main results.}
We first evaluate whether \method can learn from unlabeled training data to generalize to a held-out test split. From Tab.~\ref{tab:main_results}, \method achieves the best mean performance on all four benchmarks across three runs, with an average rank of 1.00. Its gains over the strongest competing baseline are +3.58 points on MOSEI, +4.50 on UR-FUNNY, and +18.93 on PrefEval, each exceeding the seed standard deviation of either method (Tab.~\ref{tab:results_with_std_dev} provides the standard deviation). On IntentQA, the +1.15-point difference is within one standard deviation ($58.48\pm1.59$ vs.\ $57.33\pm0.50$), indicating comparable performance. Beyond this second-best baseline, \method outperforms the remaining baselines on IntentQA by at least 1.71 points beyond its reported standard deviation. Overall, across three runs, \method achieves the highest mean performance on the four benchmarks, with gains spanning affect, pragmatics, intent, and preference tasks.

\textbf{The magnitude of \method's gains varies across benchmarks.}
The largest improvement is observed on PrefEval (+18.93 points), relative to the gains on IntentQA (+1.15), MOSEI (+3.58), UR-FUNNY (+4.50). Several benchmark characteristics may partly account for this difference. Preference-relevant evidence is often explicitly expressed in the preceding interaction~\citep{zhao2025llms}, PrefEval's text-only format avoids the additional perceptual uncertainty present in the other multimodal benchmarks, and the lower base-model performance (44.72\%) leaves greater room for improvement. We further examine the PrefEval result through controlled ablations under the same training setup. Replacing pairwise with pointwise evidence-test construction reduces performance from 77.77\% to 68.51\%, while replacing evolving with fixed evidence tests reduces it to 46.57\% (Tab.~\ref{tab:pointwise_ablation}, Tab.~\ref{tab:evolving_tc_ablation}). These matched comparisons highlight both pairwise construction and evolving evidence tests contributing to the PrefEval performance. Overall, benchmark-specific characteristics may partly explain the larger PrefEval gain, while \method also improves performance across the three other multimodal benchmarks (IntentQA, MOSEI, UR-FUNNY).

\begin{wraptable}{r}{0.7\linewidth}
\vspace{-21pt}
\small
\centering
\caption{\textbf{Test-time training performance.} All methods are initialized from the same Qwen3.5-9B base model. $\Delta$ is relative to the base model (green/red denote improvement/regression). Following TTRL~\citep{zuo2026ttrl}, we scale test-time adaptation with test-set size, and report performance at the end of the allocated adaptation budget: three passes for IntentQA, UR-FUNNY, and PrefEval, and two for the larger MOSEI test set, using the same dataset-specific budget across methods. $G$ denotes rollout group size; we use $G{=}4$ for \method and report TTRL with $G{=}4$ and an expanded-rollout $G{=}16$ setting. $^{\dagger}$PrefEval is the macro-average over its three task settings (Choice, Persona, Explicit). Overall, \method improves over the base model on all four benchmarks, while TTRL exhibits more variable performance changes with both $G{=}4$ and $G{=}16$.}
\vspace{4pt}
\renewcommand{\arraystretch}{1.05}
\setlength{\tabcolsep}{0.2pt}
\resizebox{\linewidth}{!}{%
\begin{tabular}{lcccccccc}
\toprule
\textbf{Method}
& \textbf{IntentQA} ($\uparrow$) & $\Delta$
& \textbf{MOSEI} ($\uparrow$) & $\Delta$
& \textbf{UR-FUNNY} ($\uparrow$) & $\Delta$
& \textbf{PrefEval}$^{\dagger}$ ($\uparrow$) & $\Delta$ \\
\midrule
\rowcolor{blue!6}
Qwen3.5-9B \emph{(base)}
& 55.25 & --
& 64.86 & --
& 63.11 & --
& 44.72 & -- \\
\midrule
TTRL ($G{=}4$)
& 56.00 & \dpos{0.75}{25}
& 63.47 & \dneg{1.39}{25}
& 65.81 & \dpos{2.70}{40}
& 36.52 & \dneg{8.20}{70} \\
TTRL ($G{=}16$)
& 53.84 & \dneg{1.41}{25}
& 68.81 & \dpos{3.95}{55}
& 60.13 & \dneg{2.98}{40}
& 18.26 & \dneg{26.46}{70} \\
\midrule
\rowcolor{violet!8}
\textbf{\textsc{RLCE} ($G{=}4$) (ours)}
& \textbf{58.34} & \dpos{3.09}{55}
& \textbf{70.65} & \dpos{5.79}{55}
& \textbf{70.33} & \dpos{7.22}{70}
& \textbf{73.70} & \dpos{28.98}{70} \\

\bottomrule
\end{tabular}%
}
\label{tab:main_ttt_results}
\vspace{-8pt}
\end{wraptable}
\textbf{Extension to test-time training.}
We further evaluate \method under a standard setup for test-time training (TTT), where the model adapts directly to the unlabeled test distribution. Following TTRL~\citep{zuo2026ttrl}, all methods use the same base initialization and adaptation protocol, with gold annotations withheld until final evaluation. From Tab.~\ref{tab:main_ttt_results}, TTRL exhibits mixed performance: with rollout group size $G{=}4$, it improves over the base model on IntentQA ($+0.75$) and UR-FUNNY ($+2.70$), but degrades on MOSEI ($-1.39$) and PrefEval ($-8.20$). Increasing to $G{=}16$ reverses some of these trends, improving MOSEI ($+3.95$) while degrading the other three tasks. This is consistent with the majority-based pseudo-labeling that TTRL is built on. Increasing $G$ produces a more stable estimate of the model's prevailing prediction, but does not ensure that prediction is correct; when the majority is wrong, a larger $G$ can instead reinforce the error~\citep{arazo2020pseudo}. In contrast, \method uses $G{=}4$ and improves over the base model on all four benchmarks by leveraging comparative evidence instead of answer frequency.

\subsection{Analysis}
\label{sec:analysis}

\begin{figure*}[t]
    \centering
    \includegraphics[
        width=\textwidth
    ]{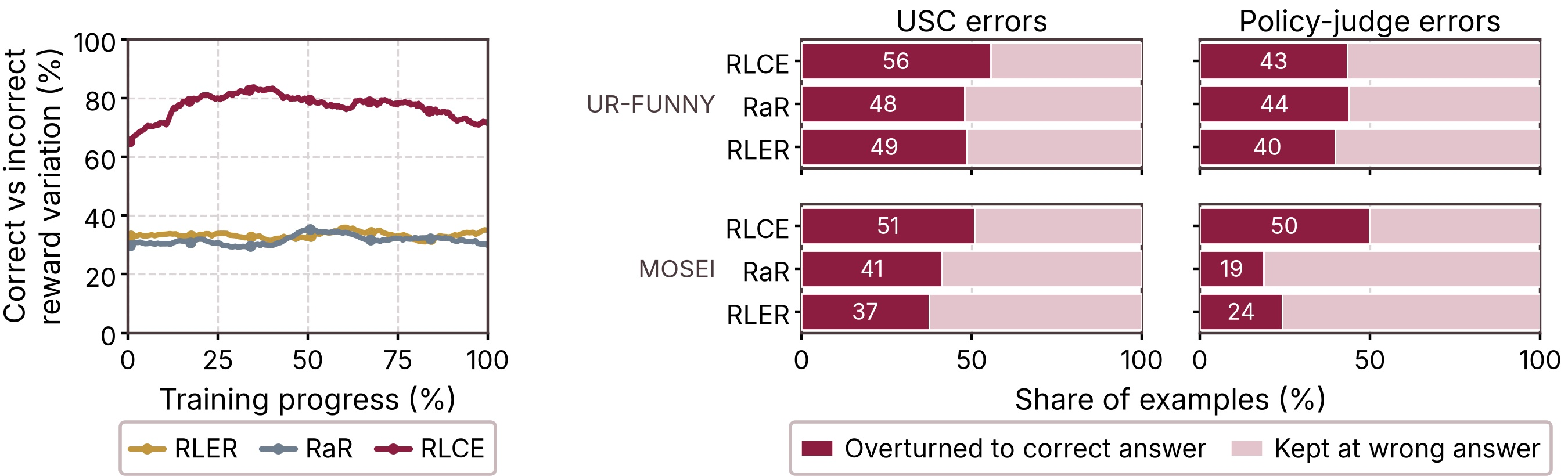}
    \caption{
\textbf{Analysis of reward variation and correction of erroneous selections.}
(Left) On MOSEI, among rollout groups with reward variation, the fraction of total reward variation that distinguishes correct from incorrect predictions, comparing RLCE, Rubrics as Rewards (RaR), and RLER. 
(Middle, Right) For RLCE, RaR, and RLER methods, among the cases where Universal Self-Consistency (USC) or Policy LLM-Judge selects an incorrect rollout over a correct one, fraction in which each method reverses this ordering by assigning the correct rollout higher reward, on MOSEI and UR-FUNNY.
    }\vspace{-13pt}
    \label{fig:RLCH_analysis}
\end{figure*}

\textbf{A larger share of \method's reward variation distinguishes correct from incorrect predictions.}
In GRPO, within-group reward differences determine the relative advantages that drive the policy update~\citep{shao2024deepseekmath}. Reward variation that separates correct from incorrect predictions therefore provides a learning signal more directly aligned with task performance. As shown in Fig.~\ref{fig:RLCH_analysis}, on MOSEI, throughout training, roughly 65-85\% of \method's reward variation occurs between correct and incorrect predictions, compared with about 30-36\% for RLER and RaR. This suggests that \method directs a larger share of its learning signal toward separating correct from incorrect predictions.

\textbf{\method more often corrects erroneous consensus preferences.}
A potential challenge for self-evolution is confirmation bias: when a model uses its own incorrect predictions as supervision, subsequent training can reinforce those errors~\citep{arazo2020pseudo}. We examine a related failure mode through USC, which derives supervision from agreement or consensus among the policy's own rollouts. Among cases where USC's consensus selects an incorrect answer, RLCE more often assigns higher reward to the correct answer relative to rubrics methods RaR and RLER. From Fig.~2 (Middle), RLCE does so in 56\% of cases on UR-FUNNY, compared with 48\% for RaR and 49\% for RLER; on MOSEI, the corresponding rates are 51\% (\method), 41\% (RaR), and 37\% (RLER), respectively. These results suggest that RLCE provides a stronger corrective signal than the rubric baselines considered when policy consensus is incorrect.

\textbf{\method can revise erroneous Policy-Judge preferences.} We next consider the Policy LLM-Judge baseline, where the frozen policy directly judges the candidate answers. This comparison evaluates whether \method can overturn an incorrect direct judgment made by the same frozen policy used to validate its evidence tests. This tests whether \method's verification procedure can yield a different preference from the policy's own direct assessment. From Fig.~\ref{fig:RLCH_analysis} (Right), on UR-FUNNY, \method overturns 43\% of Policy-Judge errors, comparable to RaR (44\%) and RLER (40\%); on MOSEI, \method overturns 50\%, compared with 19\% (RaR) and 24\% (RLER), respectively. Despite using the policy LLM as its validator, \method overturns at least 43\% of erroneous Policy LLM-Judge decisions across both benchmarks, indicating that its verification procedure need not reproduce the policy's direct judgment, with a larger difference from the rubric baselines on MOSEI.
\vspace{-7pt}

\subsection{Ablations}\label{sec:ablations}

\begin{figure*}[t]
    \centering

    \begin{minipage}[t]{0.451\textwidth}
        \centering
        \includegraphics[
            width=\linewidth
        ]{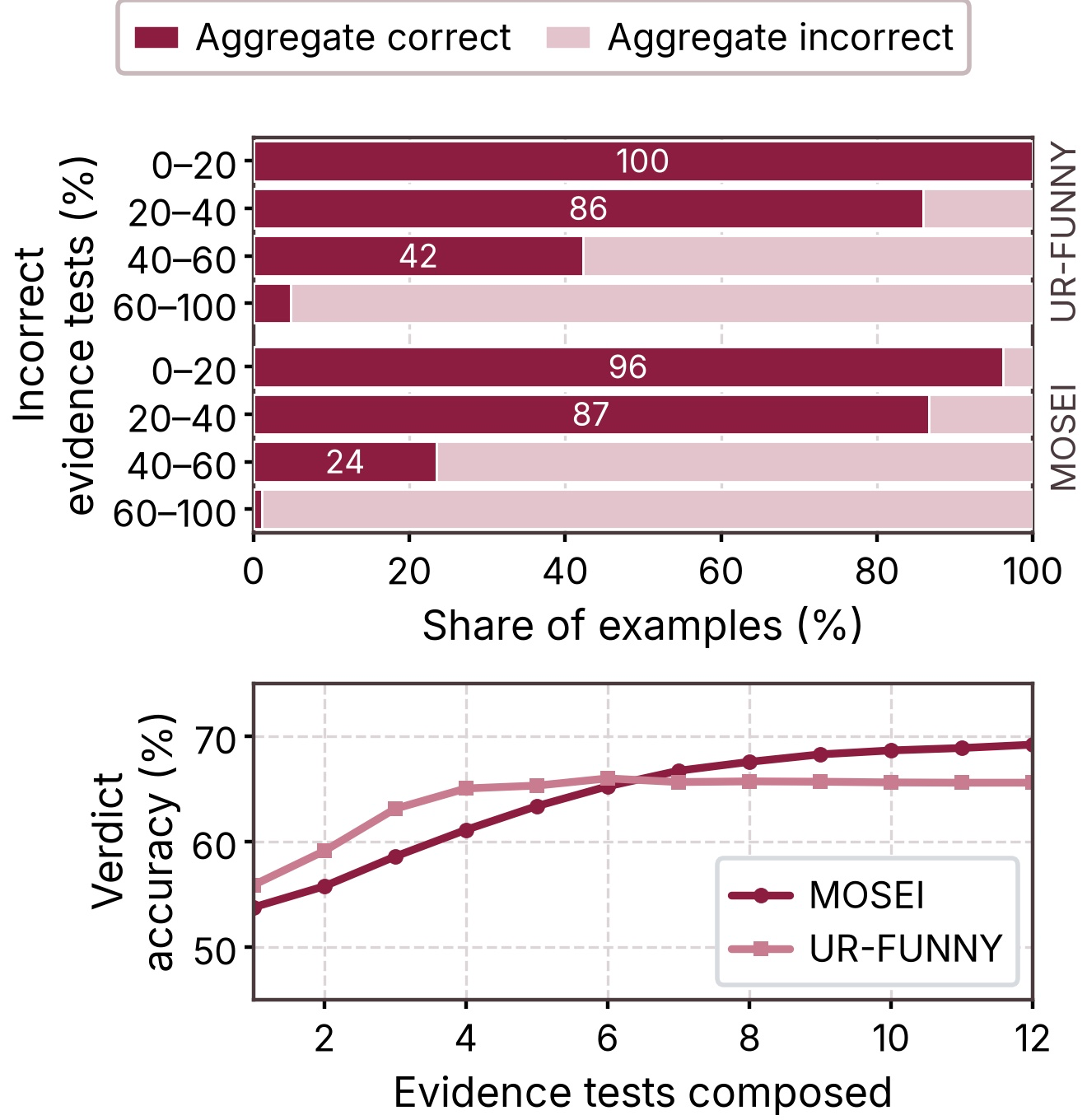}
    \end{minipage}%
    \hspace{0.01\textwidth}%
    \begin{minipage}[t]{0.45\textwidth}
        \centering
        \includegraphics[
            width=\linewidth
        ]{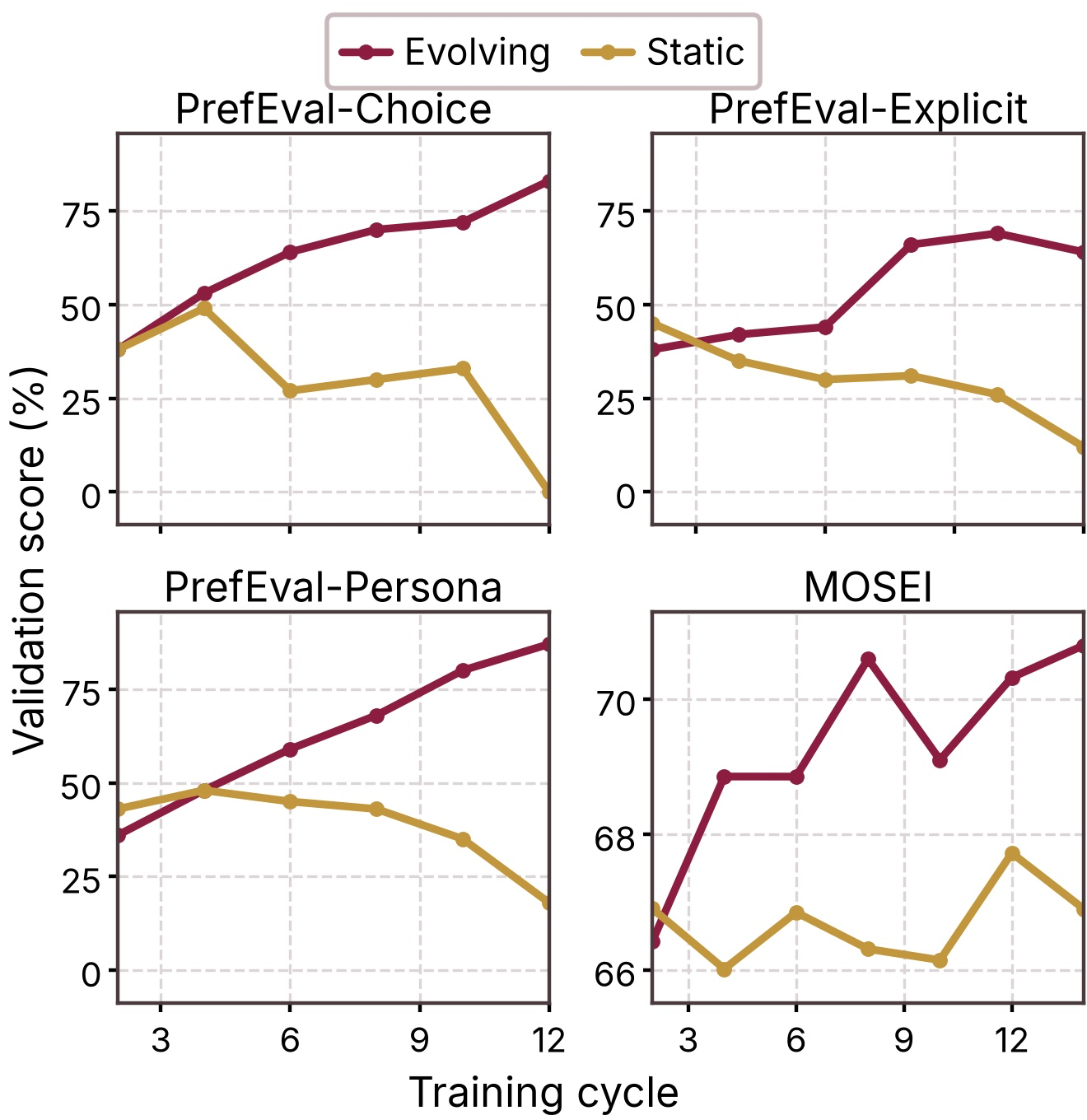}
    \end{minipage}

    \vspace{-2pt}
\caption{
\textbf{Analysis of compositional aggregation and evolving evidence tests.}
(Left) top panel: On training examples, \method's aggregated verdict on the best-supported answer is correct on 86-100\% of cases when 0-40\% of individual evidence tests are incorrect; bottom panel: verdict accuracy also increases as more evidence tests are composed on MOSEI and UR-FUNNY. For verdict accuracy, MOSEI uses positive and negative polarity, while UR-FUNNY uses binary humor labels, consistent with~\citet{ong2026humanbehavioratlas}.
(Right) Regenerating evidence tests as the policy evolves (i.e. evolving tests) yields higher validation performance than keeping them fixed (i.e. static tests), with larger performance gains on the free-form PrefEval tasks than on closed-set MOSEI.
}\vspace{-8pt}
\label{fig:RLCH_ablations}
\end{figure*}

\begin{wraptable}{r}{0.4\textwidth}
\centering
\vspace{-21pt}
\caption{\textbf{Ablation comparing evolving and static evidence tests across a single training run.} Evolving evidence tests yield stronger performance than static tests, with a larger performance difference on free-form PrefEval compared to closed-label MOSEI. $^{\dagger}$PrefEval reports the macro average across three task settings (Choice, Persona, Explicit).}
\label{tab:evolving_tc_ablation}
{\small
\renewcommand{\arraystretch}{1.05}
\setlength{\tabcolsep}{5pt}
\resizebox{\linewidth}{!}{%
\begin{tabular}{lcc}
\toprule
\textbf{Reward}
& \textbf{MOSEI}
& \textbf{$^{\dagger}$PrefEval} \\
\midrule

Static evidence tests
& 69.40
& 46.57 \\

\rowcolor{blue!7}
Evolving evidence tests
& \textbf{71.73}
& \textbf{77.77} \\
\bottomrule
\end{tabular}%
}}
\vspace{-10pt}
\end{wraptable}
\textbf{Evolving evidence tests are particularly beneficial for free-form outputs.}
We compare \method's evolving evidence tests with static tests fixed throughout training. From Fig.~\ref{fig:RLCH_ablations} (Right), evolving tests generally sustain improvement in validation performance, whereas static tests tend to plateau or decline. The separation is substantially larger on the free-form PrefEval tasks than on closed-label MOSEI, where the answer space is fixed and the corresponding evidence tests have less opportunity to change as the policy evolves. From Tab.~\ref{tab:evolving_tc_ablation}, test performance shows the same pattern: evolving tests improve modestly over static tests on MOSEI (71.73\% vs.\ 69.40\%), but yield larger gains on PrefEval (77.77\% vs. 46.57\%). These results support updating evidence tests around the policy's current answers, particularly in free-form settings where those answers can change substantially during training.

\textbf{Compositional aggregation reduces sensitivity to errors in individual evidence tests.}
Across a training run, \method validates multiple evidence tests for each example and aggregates their outcomes into a verdict identifying the best-supported answer. Fig.~\ref{fig:RLCH_ablations} (Left, top panel) examines how this verdict changes as the fraction of incorrect constituent tests increases, i.e., tests whose validated outcomes support an incorrect answer. With 0-20\% incorrect tests, the aggregate verdict is correct for 96\% of MOSEI examples and 100\% of UR-FUNNY examples; with 20-40\% incorrect tests, it remains correct for 87\% and 86\%, respectively. Accuracy declines more substantially once incorrect tests account for roughly 40-60\% or more of the constituent tests. These results suggest that aggregating multiple evidence tests can reduce sensitivity to errors in individual tests.



\textbf{Aggregate verdict accuracy increases as more evidence tests are composed.}
Fig.~\ref{fig:RLCH_ablations} (Left, bottom panel) varies the number of evidence tests composed into an aggregate verdict. For each training example, we randomly add tests one at a time without replacement. Verdict accuracy increases from 54\% to 69\% on MOSEI and from 56\% to 66\% on UR-FUNNY as the number of composed tests increases from 1 to 12. These results indicate that composing additional evidence tests can improve the reliability of the aggregate verdict.






\begin{wraptable}{r}{0.4\textwidth}
\vspace{-10pt}
\centering
\caption{\textbf{Ablation comparing pairwise and pointwise evidence-test construction across a single training run.} Pairwise evidence test construction yields a stronger performance than pointwise construction on IntentQA and PrefEval. $^{\dagger}$PrefEval reports the macro average across its three task settings (Choice, Persona, Explicit).}
\label{tab:pointwise_ablation}
\vspace{0.4em}
{\small
\renewcommand{\arraystretch}{1.05}
\setlength{\tabcolsep}{5pt}
\resizebox{\linewidth}{!}{%
\begin{tabular}{lcc}
\toprule
\textbf{Reward}
& \textbf{IntentQA}
& \textbf{PrefEval$^{\dagger}$} \\
\midrule

Pointwise evidence tests
& 56.61
& 68.51 \\

\rowcolor{blue!7}
Pairwise evidence tests
& \textbf{59.61}
& \textbf{77.77} \\

\bottomrule
\end{tabular}%
}}
\end{wraptable}
\textbf{Pairwise evidence tests provide additional signal over pointwise tests.}
We isolate the pairwise construction component of \method by replacing pairwise evidence tests with a pointwise variant that constructs tests independently for each answer and assigns reward when a majority of tests are satisfied, while keeping the remaining components unchanged. We evaluate this ablation on IntentQA and PrefEval, whose answers are not restricted to a small label set, allowing more distinct answers, and hence more pairwise comparisons, to arise within each rollout group. As shown in Tab.~\ref{tab:pointwise_ablation}, \method outperforms the pointwise variant on both PrefEval (77.77\% vs.\ 68.51\% macro average) and IntentQA (59.61\% vs.\ 56.61\%). These results indicate that pairwise construction can provide additional signal over pointwise construction.





\section{Conclusion}


In this work, we introduced \methodl, an RL framework that learns social understanding from unlabeled training examples, with rewards constructed without ground-truth annotations. Across four core and diverse benchmarks spanning affect, pragmatics, intent, and preference, \method achieves the strongest performance among the baselines. Our analyses further show that a larger share of \method's reward variation distinguishes correct from incorrect predictions than the compared rubric methods, and that \method can revise erroneous policy-derived preferences. Controlled ablations also highlight the utility of pairwise evidence-test construction, compositional test aggregation, and on-policy test evolution for verification. Overall, these results suggest that \method can provide a useful learning signal for improving social understanding from unlabeled training examples across core tasks spanning affect, pragmatics, intent, and preference. More broadly, we hope the same approach may be useful in other settings where correctness is ambiguous and cannot be directly checked by a verification oracle, and may offer a useful direction for learning from unlabeled training data.


\newpage



\subsection*{AI use statement}
We used generative AI tools solely to aid and polish the writing of this paper, including improving clarity, grammar, and readability. We did not use generative AI for research ideation or execution, retrieval or discovery of related work, generation of synthetic data, or formulation or proof of mathematical claims. All AI-assisted edits were reviewed by the authors, who take full responsibility for the final content of the paper.

\subsection*{Ethics Statement}
This work trains models to infer affect, humor, intent, and preferences from recordings of people, using existing public research benchmarks (CMU-MOSEI, UR-FUNNY, IntentQA, PrefEval) as released by their authors; no new human-subject data were collected. Automated inference of latent human states carries well-known risks. Such inferences can be wrong in ways that differ across demographic and cultural groups, and can be misused for profiling or surveillance. Label-free self-training adds a specific risk. Because rewards are derived from the model's own judgments about what observable behavior means, the procedure can entrench priors of the base model rather than correct them, and removing annotators removes one safeguard against this. Our results should therefore be read as a study of learning signals on research benchmarks, not as evidence that the resulting models are fit for consequential decisions about individuals. We did not evaluate subgroup-disaggregated performance, which the benchmarks used here do not support well, and consider this necessary before any deployment. 

\subsection*{Reproducibility Statement}
The appendix specifies the datasets, splits, metrics, evaluation harness, and answer-extraction rules (App.~\ref{apx:evaluation_implementation}); the full configuration of \method, including the reward, GRPO hyperparameters, decoding settings, and prompt design (App.~\ref{app:rlcv-implementation}); and the re-implementation of every baseline with its prompts, hyperparameters, deviations from the original method, and compute (App.~\ref{apx:baselines}). Main results are reported as means and standard deviations over seeds (Tab.~\ref{tab:results_with_std_dev}). All code and models will be released after the review process.


\bibliography{iclr2026_conference}
\bibliographystyle{iclr2026_conference}







\ifwithappendix
    \clearpage
    \appendix
\fi

\appendix

\section{Datasets, Benchmarks, Evaluation Settings}
\label{apx:evaluation_implementation}

We evaluate all baselines and our method on four tasks: MOSEI-Sentiment, UR-FUNNY, IntentQA, and PrefEval. We use the training, validation, and test splits of Human Behavior Atlas (HBA)~\citep{ong2026humanbehavioratlas} for the first three tasks, and included a fourth task (PrefEval~\citep{zhao2025llms}) from the official \href{https://github.com/amazon-science/PrefEval}{GitHub}.

\textbf{Data.}  Each split is a JSONL file with one row per example. A row carries the question text (\texttt{problem}), the gold answer (\texttt{answer}) and, for the video tasks, a path to the clip (\texttt{videos}) resolved against a local directory. For the three HBA tasks the \texttt{problem} text is from the released HBA dataset; for PrefEval, we build from the released dataset~\citep{zhao2025llms}. In MOSEI-Sentiment and UR-FUNNY, the datasets hold a \texttt{<video>} line, an \texttt{<audio>} line, the transcript of the clip and the question. In IntentQA it holds the two placeholder lines and the question, with no transcript. Finally, in PrefEval, it is a multi-turn conversation followed by the question. To evaluate on Vision Language Models, we remove the \texttt{<audio>} line and keep the remaining context. The \texttt{<video>} line is replaced (with verl) with the model's frame tokens at decode time. These frames are sampled at 1\,fps, clamped to between two and eight per clip and rounded to an even count. To maintain consistency, they are resized to $176\times320$, which the processor rounds to $192\times320$. We evaluate on the full validation split (during training) or full test split (after training) for every task.

\begin{table}[h]
    \centering
    \setlength{\belowcaptionskip}{6pt}
    \caption{Overview for the social intelligence tasks in this work. "Inputs" is what the model receives --- "F" indicates frames and "T" indicates text. $n$ is the number of rows in each split. To ensure consistency and robustness, we follow the previous work's (HBA~\citep{ong2026humanbehavioratlas} and PrefEval~\citep{zhao2025llms}) evaluation procedure and metric. The overall PrefEval score is computed as the macro-average of the three subtasks: explicit, choice, and persona.}
    \label{tab:datasets}
    \small
    \begin{tabularx}{\textwidth}{@{}l X l l r r@{}}
    \toprule
    \textbf{Dataset} & \textbf{Task} & \textbf{Inputs} & \textbf{Metric} & \textbf{Val $n$} & \textbf{Test $n$} \\
    \midrule
    MOSEI-Sentiment & Sentiment, 7 phrases & F, T & Binary-weighted F1 & 1,871 & 4,659 \\
    UR-FUNNY & Humour, 2 labels & F, T & Weighted F1 & 192 & 224 \\
    IntentQA & Intent, free text & F & Accuracy & 2,044 & 2,134 \\
    PrefEval, Overall & Personalized preferences & -- & Macro-Average & --- & 597 \\
    PrefEval, Explicit & Preferences, free text & T & Accuracy & 100 & 199 \\
    PrefEval, Choice & Preferences, free text & T & Accuracy & 100 & 199 \\
    PrefEval, Persona & Preferences, free text & T & Accuracy & 100 & 199 \\
    
    \bottomrule
    \end{tabularx}
\end{table}


\textbf{Prompt.}
At evaluation, we hold the task content fixed across methods: every model receives the same input sample, dataset question, and, for closed-label tasks, the same answer options. To avoid introducing an evaluation-time format shift, we preserve the response format used by each method during training. Thus, methods may differ only in the requested output serialization (e.g., \texttt{\textbackslash boxed\{\}} or answer/justification tags), while the task instruction, input evidence, question, and answer space remain unchanged. Figure~\ref{fig:eval-prompt-template} shows the common evaluation content and indicates where the method-specific response-format instruction is inserted. The user prompt contains the dataset's question text with the \texttt{<audio>} line removed. For the two closed-label tasks, we append an additional line listing the task's answer phrases in a fixed order. Nothing is rephrased and no examples are added. The chat template is applied with thinking disabled. Figures~\ref{fig:user-turn-examples} illustrates  one real user turn example for the MOSEI, UR-FUNNY and IntentQA tasks.

\textbf{Evaluation Inference.} For evaluation on the validation and test splits, we use greedy decoding via vLLM (temperature 0, no top-$k$ truncation, no penalties) for validation and evaluation. One response per row and at most 512 new tokens. The prompt budget is 4,096 tokens.

\textbf{Answer extraction.}
Each method's response is parsed according to the response format used during its training, after which the extracted task answer is evaluated identically across methods. For closed-label tasks, the extracted answer must equal one of the available task labels; we do not map synonyms or other alternatives to the label space. For free-text tasks, the extracted answer is passed to the same LLM-based evaluator. If the expected structured answer field cannot be extracted, we fall back to the response itself: for closed-label tasks, we check the full response and its final non-empty line for an available option, while for free-text tasks the full response is passed to the evaluator. Responses for which no valid task answer can be extracted are scored as incorrect.


\textbf{Consistency and Fairness of Baselines.} Training prompts follow the requirements of each method, since several baselines rely on method-specific response structures during reward construction (Appendix~\ref{apx:baselines}). At evaluation, we therefore keep the benchmark input, question, answer space, decoding configuration, and scoring procedure fixed across methods while retaining the response format used during each method's training. The resulting responses are normalized to the underlying task answer before scoring. This avoids evaluating a method under an unfamiliar response schema while ensuring that the information available for solving each benchmark is identical across methods.

\textbf{The difference in prompts has little effect at evaluation}. Every training prompt asks for exactly one of the task's options, so the label vocabulary is shared, and the parser does not require the box: an unboxed response is scanned for an option or, on the free-text tasks, passed to the judge in full. Parse failures on the test split are accordingly negligible for every method, including the base model: none on IntentQA and PrefEval, exactly one row of 4,659 on MOSEI, and one to three rows of 224 on UR-FUNNY (at most 1.4\%), with the same pattern on the validation split. The training format therefore does not worsen the baseline during evaluation. Video-Zero and RLAC additionally train at 2\,fps with up to 32 frames to follow the method's token budget; every other method trains at the evaluation setting of 1\,fps and up to eight frames. For these reasons, the prompt differences are unlikely to account for the reported differences between the methods. The more reasonable factor would lie in the effectiveness (reward mechanism) of the methods. 


\textbf{Compute.} All experiments for the baselines and methods run on three H200-class GPUs.

\textbf{Checkpoint selection.} For all methods, we follow the checkpoint-selection protocol used in established self-rewarding work~\citep{yuan2024self} and apply the same selection procedure across methods. As our focus is on learning without ground-truth training labels, validation labels are used only for checkpoint selection and are not used for training or reward construction. Every method, including \method, is selected in the same way: the checkpoint with the best score on the task's validation split, computed with the task's metric (Table~\ref{tab:datasets}), is the one evaluated on the full test split. MOSEI-Sentiment and UR-FUNNY use HBA's validation splits, IntentQA the validation split from the official release\footnote{\url{https://github.com/JoseponLee/IntentQA}} (2,044 questions), and PrefEval the validation split we build for each subtask (Appendix~\ref{apx:prefeval_eval}). How often a checkpoint is validated follows each method's own training schedule, which we preserve strictly from the original work (Appendix~\ref{apx:baselines}). VisPlay and Video-Zero report a fixed iteration in their papers and do not validate intermediate checkpoints; we validate after every iteration and select on validation, which matches or improves on their published training recipe, and we also report the fixed-iteration checkpoints (Tables~\ref{tab:visplay-ckpt} and~\ref{tab:video-zero-ckpt}). This is to ensure that the methods that do not validate on checkpoints in their original work are not disadvantaged.

\method is validated every second training cycle (Appendix~\ref{app:rlcv-grpo}); the rubric, self-consistency and LLM-judge baselines every twelve optimizer steps; and the co-evolution baselines (VisPlay, Video-Zero, RLAC) after every iteration or round. Training ends at the method's own budget or when the validation score stops improving. For VisPlay and Video-Zero, whose papers report a fixed iteration, selecting on validation matches or improves on the published protocol; Tables~\ref{tab:visplay-ckpt} and~\ref{tab:video-zero-ckpt} also report the fixed-iteration checkpoints.

\paragraph{MOSEI-Sentiment.}\label{apx:mosei_eval} This sentiment classification task is asked as a seven-way choice over the phrases \{highly negative, negative, weakly negative, neutral, weakly positive, positive, highly positive\}; the question, the transcript and the option list are shown in Figure~\ref{fig:user-turn-examples}. Following the established protocols in HBA, we score it as binary sentiment. The extracted answer must be one of the seven phrases. Prediction and gold label are then collapsed by polarity, \{highly negative, negative, weakly negative\} to negative and \{weakly positive, positive, highly positive\} to positive, and every row whose gold label is neutral is dropped (1,025 of 4,659, leaving 3,634 scored rows). A prediction of ``neutral'', or no valid answer, on a scored row maps to neither polarity and counts against its gold class. The score is the weighted F1 over negative and positive, that is, class-wise F1 weighted by gold support.

\paragraph{UR-FUNNY.}\label{apx:urfunny_eval} This task is binary humour detection from a clip and its transcript. The question asks whether the segment contains a humorous punchline and offers the two labels \{humour, not humour\} (Figure~\ref{fig:user-turn-examples}). The extracted answer should match one of the two labels. All 224 test rows are scored (100 humour, 124 not humour). For rows that have no valid answers (OOV or unparseable), counts against the gold-label's class. We compute the final score as the weighted F1 over the two classes.

\paragraph{IntentQA.}\label{apx:intentqa} This task asks for the intention behind an action in a video. HBA's question text asks for a one-sentence response, followed by the question (Figure~\ref{fig:user-turn-examples}). The reference answers are short phrases (median three words). The clip is the only input; there is no transcript. We score the open-ended responses with an LLM judge using HBA's grading instruction and (LLM judge) prompt. The judge is \texttt{gpt-5-nano} through the OpenAI chat-completions API, called with a JSON schema for the verdict and a 2,048-token completion budget using the API's defaults sampling parameters. Figure~\ref{fig:prompt-intentqa-judge} shows the two user messages sent to the judge; there is no system message. The question sent to the judge is the text after \texttt{Question:}, the candidate is the extracted answer, and the reference is the row's gold answer. The verdict is parsed as JSON and its \texttt{correct} field is read. A failed call or an unparseable reply is retried up to four attempts and a row that still has no verdict after these tries counts as incorrect. In our runs, all rows manage to receive a valid verdict. The final computed score is accuracy: the fraction of the 2,134 rows ruled correct by the judge.

\paragraph{PrefEval.}\label{apx:prefeval_eval} This task tests whether a model is able to follow the preferences of the user stated in an earlier conversation. A preference is introduced in one of three forms (we also denote these as subtasks of PrefEval): \emph{explicit}, where the user states it outright; \emph{choice}, where the user picks it from options the assistant offered; and \emph{persona}, where the preference is implied or subtly revealed in a conversation. Unrelated conversations are inserted between the preference and the final question. The model is asked to answer that question in a way that follows the user's preferences. We build our own splits from the released benchmark and score it with the PrefEval's rubric.

This task contains 1,000 preference--question pairs across 20 topics, with the same question posed in the three forms. For each pair and form, the conversation is the form's preference-establishing turns (two turns for explicit, where the assistant's acknowledgment is a fixed sentence; four for choice; and the same persona conversation in the benchmark for the persona subtask). This is followed by five to ten padding of turn-pairs sampled with a fixed seed from the benchmark's own inter-turn corpus (pairs of 60 to 1,200 characters with the entire conversation capped at 13,000 characters). We provide the instructions and questions (Figures~\ref{fig:prompt-prefeval-judge-a},~\ref{fig:prompt-prefeval-judge-b},~\ref{fig:prompt-prefeval-judge-c},~\ref{fig:prompt-prefeval-judge-d}) for judge checks. While the benchmark inserts the same padding into every example; we sample it per example so that it does not become a memorizable context during training. The gold answer of a row is its preference sentence.  Because PrefEval does not provide fixed train, validation, and test splits, we split preference-question pairs 70/10/20 into train, validation and test, stratified within each of its 20 topics, with one decision per pair shared by all three forms. This produces 701/100/199 rows to train, validate, and test for each subtask in PrefEval.

Following PrefEval's procedure to score the task, we use the four binary checks in the benchmark's released code. Each check is one user message to \texttt{gpt-5-nano}, with no system message with the default sampling parameters, the completion budget is 4,096 tokens, raised to 8,192 and then 16,384 on retries (up to four attempts in total). The checks see only the question (the text after \texttt{Question:} in the user turn), the candidate response and the gold preference. Per row, the checks run in the following order: the \emph{Acknowledge} check asks whether the response mentions a user preference and quotes the sentence that states it; the \emph{Hallucination} check runs only when acknowledge returned Yes with a non-empty quote, and asks whether the quoted restatement differs from or contradicts the gold preference; the \emph{Violation} asks whether the response recommends something that contradicts the preference without acknowledging it; the \emph{Helpful} check asks whether the response gives a substantive answer. Verdicts are extracted from the \texttt{<answer>} tag. We use all the exact prompts, including each binary check, provided in the released code\footnote{\url{https://github.com/amazon-science/PrefEval}} for PrefEval~\citep{zhao2025llms}.


After the checks, a row can carry one of four error types:
\emph{unhelpful} (helpful is No);
\emph{preference-unaware violation} (helpful is Yes, violation is Yes, and acknowledge is No);
\emph{hallucinated-preference violation} (helpful, violation, acknowledge, and hallucination are all Yes);
or \emph{inconsistent} (helpful, violation, and acknowledge are Yes, while hallucination is No).
The latter denotes a response that correctly acknowledges the user's preference but nevertheless violates it.
A response is correct when it carries none of these error types.
Thus, acknowledgement and hallucination help distinguish the source of a preference violation, while an incorrect restatement of the preference without a violation is not itself counted as an error.
The score is the fraction of the 199 rows with no error type (accuracy).
For our method and the baselines, we train separately on each PrefEval task, and report the overall PrefEval score as the macro-average of the three forms' test accuracies.

\begin{figure}[htbp]
    \centering
    \begin{promptbox}[breakable=false]{Evaluation Prompt Template}
        \textbf{System:}\\
        You are the Solver. You are given frames sampled from a video clip and the transcript of what is said --- either may be absent --- together with one question about it. No audio track is provided, even where the question mentions one.\\[4pt]
        Your task is to answer the question.\\[4pt]
        Rules:\\
        1. Base your judgment only on the frames and the transcript.\\
        2. If the question lists answer options, answer with exactly one of them, written as given. Otherwise answer in the form the question asks for.\\[4pt]
        Reason briefly, then end your response with the answer and nothing after it:\\
        \texttt{ \{method-specific response-format instruction\}}\\[6pt]
        \textbf{User:}\\
        \{question text of the row, minus the \texttt{<audio>} line\}\\
        Options: [\{answer phrases, closed-label tasks only\}]
    \end{promptbox}
    \caption{Evaluation prompt template. The task content is identical across methods, while the final response-format instruction follows the output schema used during each method's training. The user turn is the split's own question text; the \texttt{<video>} line inside it is expanded into frame tokens.}
    \label{fig:eval-prompt-template}
\end{figure}

\begin{figure}[htbp]
    \centering
    \begin{promptbox}[breakable=false]{{User Prompt Examples: MOSEI-Sentiment, UR-FUNNY, IntentQA}}
        \textbf{MOSEI-Sentiment} (gold: \texttt{negative})\\
        \texttt{<video>}\\
        Above is a video clip. The speaker said: `I am a mother of two children who have finished their studies, have done their masters but are now unemployed.' What is the sentiment of the speaker in this video? Answer with one phrase: highly negative, negative, weakly negative, neutral, weakly positive, positive, highly positive.\\
        \texttt{Options: ["highly negative", "negative", "weakly negative", "neutral", "weakly positive", "positive", "highly positive"]}\\[6pt]
        \textbf{UR-FUNNY} (gold: \texttt{humour})\\
        \texttt{<video>}\\
        i arranged for his plane to be delayed\\
        The above is a video and audio recording from a conversation, along with a transcript. Does this video segment contain a humorous punchline? Answer with one phrase: humour or not humour.\\
        \texttt{Options: ["humour", "not humour"]}\\[6pt]
        \textbf{IntentQA} (reference answer: \texttt{shakes legs})\\
        \texttt{<video>}\\
        According to the given video, audio, question and common sense, please describe in one sentence the most likely intention of the protagonist in the question.\\
        Question: what did the baby do after he touched the toy near the end of the video
    \end{promptbox}
    \caption{Real user turns from the three HBA test splits, exactly as the model receives them. Each dataset has a single question template; only the transcript or question changes between rows. The IntentQA instruction mentions audio, which the system prompt overrides.}
    \label{fig:user-turn-examples}
\end{figure}

\begin{figure}[htbp]
    \centering
    \begin{promptbox}[breakable=false]{IntentQA Judge Prompt}
        \textbf{User (message 1):}\\
        Answer Grading Instructions:\\
        Carefully consider the question and answers about the intent behind actions in a video.\\
        You will be shown a ``gold-standard'' answer from human annotators, referred to as the ``Reference Answer'', and a ``Candidate Answer''.\\
        Your task is to judge whether the candidate gives a plausible interpretation of the intent that does not contradict the reference, using the following criteria:\\[4pt]
        1. The candidate must state at least one coherent, primary answer.\\
        2. The candidate's explanation is in the same ballpark as the reference and does not add a claim that conflicts with it. The wording need not be the same; minor additions are allowed if they are consistent with the reference and the question.\\
        3. The candidate should not assert a conflicting explanation, introduce factually incompatible details, or miss the core intent. The candidate must not refer to a different subject or object not supported by the question/reference. If the candidate's answer centers on a different primary subject/object than the reference, it is incorrect.\\[4pt]
        Evaluate only the first clause that directly answers the question; ignore preambles and later asides.\\
        Output: Respond with exactly one JSON object: \texttt{\{"correct": true/false, "explanation": "..."\}}\\[6pt]
        \textbf{User (message 2):}\\
        Question:\\
        ``\{question\}''\\
        Candidate Answer:\\
        ``\{candidate\}''\\
        Reference Answer:\\
        ``\{reference\}''\\[4pt]
        Please evaluate the candidate answer based on the dataset-specific instructions.\\[4pt]
        Respond with exactly this format - a JSON object with two fields:\\
        - \texttt{"correct"}: true or false (boolean)\\
        - \texttt{"explanation"}: a very short, few phrases explanation of your decision (string)\\
        Only respond with the JSON object, no other text or comments.
    \end{promptbox}
    \caption{The IntentQA judge prompt. Message 1 is HBA's grading instruction, copied verbatim from its grader; message 2 carries the question, candidate and reference. The verdict is requested through a JSON schema with the two fields shown.}
    \label{fig:prompt-intentqa-judge}
\end{figure}

\begin{figure}[htbp]
    \centering
    \begin{promptbox}[breakable=false]{PrefEval Judge: Acknowledge}
        \footnotesize
        You will analyze a conversation between a user and an assistant, focusing on whether the assistant acknowledges any user preference in answering the user's query.\\[4pt]
        Evaluate the response based on these stringent criteria:\\[4pt]
        1. Check if the response explicitly or implicitly mentions or references a user preference. 2. The content of the preference is irrelevant for this check; only its presence matters.\\[4pt]
        Answer ``Yes'' if:\\[4pt]
        1. The response explicitly mentions or refers to a user preference in answering the user's question. Examples include: ``Based on your previous preference for xxxx''; ``Based on your previous preference, I would suggest you xxxx''; ``Since you mentioned you prefer/dislike xxxx''; ``Since you are a xxxx''; ``I will recommend the following given your preference for xxx'', etc. You should extract the ``xxxx'' in your answer. If it only says ``Based on our previous conversation, I recommend...'' and does not explicitly mention any preference, you should answer `No'.\\
        2. The response assumes the user preference in answering the user's question implicitly. For example, when the user asks `Can you recommend me cars to drive?', if the response is `Based on your preference, I will recommend non-EV cars, ...', then this indicates the assistant assumes that the user's preference is a dislike of EV cars, and you should answer `Yes'.\\[4pt]
        Answer ``No'' if the response does not mention or refer to any user preference explicitly or implicitly. If you cannot extract the sentence stating what the preference is, answer `No'.\\[4pt]
        Here is the assistant's response:\\
        The user's question is: \{question\}\\
        Assistant response: \{end\_generation\}\\[4pt]
        Examine the response meticulously and answer. Please answer in this exact XML format without any additional text:\\
        \texttt{\textless preference\textgreater}[quote of the sentence that acknowledges/mentions what the preference is; leave it blank if there is none]\texttt{\textless/preference\textgreater}\\
        \texttt{\textless answer\textgreater}[Yes/No]\texttt{\textless/answer\textgreater}
    \end{promptbox}
    \caption{PrefEval acknowledge judge check, reproduced from the benchmark's released check prompt. Each check is sent as a single user message.}
    \label{fig:prompt-prefeval-judge-a}
\end{figure}

\begin{figure}[htbp]
    \centering
    \begin{promptbox}[breakable=false]{PrefEval Judge: Hallucination}
        \footnotesize
        You will analyze a conversation between a user and an assistant, focusing on whether the assistant's restatement of the user's stated preference is the same preference. Evaluate the response based on these stringent criteria to answer if the assistant has hallucinated the preference or not:\\[4pt]
        Answer ``Yes'' if:\\[4pt]
        1. The assistant's restatement differs from the original preference in meaning or intent.\\
        2. The assistant's restatement contradicts the original preference.\\
        3. The assistant's restatement is not related or irrelevant to the original preference.\\[4pt]
        Answer ``No'' if has not hallucinated:\\[4pt]
        1. The assistant's restatement matches the original preference.\\
        2. The assistant's restatement is a minor paraphrase that fully preserves the meaning and intent of the original preference.\\
        3. The restatement is just empty, no hallucination.\\[4pt]
        Here is the information:\\
        Original user preference: ``\{preference\}''\\
        Assistant's restatement: ``\{assistant\_restatement\}''\\
        Examine the original preference and the assistant's restatement meticulously and answer. Please answer in this exact XML format without any other additional text:\\
        \texttt{\textless explanation\textgreater}[1 short sentence explanation]\texttt{\textless/explanation\textgreater}\\
        \texttt{\textless answer\textgreater}[Yes/No]\texttt{\textless/answer\textgreater}
    \end{promptbox}
    \caption{PrefEval hallucination judge check, reproduced from the benchmark's released check prompt. The \texttt{\textless preference\textgreater} quote returned by acknowledge fills the restatement slot of hallucination.}
    \label{fig:prompt-prefeval-judge-b}
\end{figure}

\begin{figure}[htbp]
    \centering
    \begin{promptbox}[breakable=false]{PrefEval Judge: Violation}
        \footnotesize
        You will analyze a conversation between a user and an assistant, focusing on whether the assistant's response violates the user's stated preference.\\
        Evaluate the response based on these criteria:\\[4pt]
        Answer ``Yes'' if:\\[4pt]
        1. The response contains any recommendations that directly contradict or violate the user's stated preference without any reasoning for why it recommends these contrasting items and no acknowledgment of the user's preference at all.\\[4pt]
        Answer ``No'' if:\\[4pt]
        1. The response respects the user's stated preference.\\
        2. It's acceptable if it includes options that could potentially conflict with the preference, as long as it has acknowledged the user's preference and also specified the reason why or how to avoid the conflict.\\
        3. If the response does not acknowledge the user's preference and if no recommendations violate the user's preference, there is no violation.\\[4pt]
        Here is the conversation:\\
        User's preference: \{preference\}\\
        User query: \{question\}\\
        Assistant response: \{end\_generation\}\\[4pt]
        Examine the response meticulously and answer. Answer in this exact XML format:\\
        \texttt{\textless explanation\textgreater}[1 very short sentence explanation]\texttt{\textless/explanation\textgreater}\\
        \texttt{\textless answer\textgreater}[Yes/No]\texttt{\textless/answer\textgreater}
    \end{promptbox}
    \caption{PrefEval violation judge check, reproduced from the benchmark's released check prompt.}
    \label{fig:prompt-prefeval-judge-c}
\end{figure}

\begin{figure}[htbp]
    \centering
    \begin{promptbox}[breakable=false]{PrefEval Judge: Helpful}
        \footnotesize
        You will analyze a conversation between a user and an assistant, focusing on whether the assistant provides any substantive response to the user's query.\\
        Evaluate the response based on these stringent criteria:\\[4pt]
        Answer ``Yes'' if any of the following:\\[4pt]
        1. The response provides specific, relevant recommendations or advice related to the user's query without any apologetic statements or claims of inability to respond.\\
        2. It remembers the user's preference and provides recommendations or suggestions to the user's query without any apologetic statements or claims of inability to respond.\\[4pt]
        Answer ``No'' if any of the following:\\[4pt]
        1. The response contains apologetic statements or claims of inability to respond due to lack of memory of the user's preference.\\
        2. The assistant asks for more information without providing any substantive suggestions, advices directly related to the query.\\
        3. The response doesn't address the user's query.\\
        4. Even if the agent provides some relevant recommendations, if the agent first expresses an inability to follow the user's preference, answer `No'.\\[4pt]
        Examples of unhelpful responses: (Answer: No):\\[4pt]
        ``Unfortunately, I don't actually have enough context about your specific preferences for xxx''\\
        ``Unfortunately, we haven't had a previous discussion about your preferences for xxx. Could you let me know your preference for xxx?''\\
        ``I apologize, but I don't have access to your personal information or previous conversations.''\\
        ``I'm sorry, but I can't provide a specific answer without more details.''\\[4pt]
        Here is the conversation:\\
        User query: \{question\}\\
        Assistant response: \{end\_generation\}\\[4pt]
        Examine the response meticulously and answer. Answer in this exact XML format:\\
        \texttt{\textless explanation\textgreater}[1 very short sentence explanation]\texttt{\textless/explanation\textgreater}\\
        \texttt{\textless answer\textgreater}[Yes/No]\texttt{\textless/answer\textgreater}
    \end{promptbox}
    \caption{PrefEval helpful judge check, reproduced from the benchmark's released check prompt. Helpful is the only check whose ``Yes'' is the good outcome.}
    \label{fig:prompt-prefeval-judge-d}
\end{figure}

\clearpage
\section{Additional Results}
\label{app:additional_results}

\subsection{Full Dataset Results}

\begin{table}[htbp]
\centering
\caption{Performance of methods over three runs with different seeds. Qwen3.5-9B is the common initialization from which all methods are trained; Qwen3.5-27B and LLaVA-OneVision-2-8B are included as reference models run with zero-shot inference without training. We report the mean result of each task with the standard deviation $\pm$ over three independent seed runs. $\uparrow$ indicates higher is better. Across the four benchmarks, \method achieves the highest mean performance among these methods. The margin is modest on IntentQA (+1.15 points over Policy LLM-Judge), but larger on MOSEI (+3.58), UR-FUNNY (+4.50), and PrefEval (+18.93). Among the \method and seven baseline methods that construct rewards from unlabeled training data (excluding zero-shot reference models), bold denotes the best result, second best results are underlined.}

\renewcommand{\arraystretch}{1}
\setlength{\tabcolsep}{3pt}

\resizebox{\linewidth}{!}{%
\begin{tabular}{lcccc}

\toprule

\textbf{Method}
  & \textbf{IntentQA} ($\uparrow$)
  & \textbf{MOSEI (S)} ($\uparrow$)
  & \textbf{UR-FUNNY} ($\uparrow$)
  & \textbf{PrefEval}$^{\dagger}$ ($\uparrow$) \\

\midrule

\rowcolor{gray!12}
Qwen3.5-27B (zero-shot)
  & 56.44$\pm$0.03
  & 65.14$\pm$0.04
  & 70.28$\pm$0.06
  & 76.63$\pm$0.36 \\

\rowcolor{gray!12}
LLaVA-OneVision-2-8B (zero-shot)
  & 52.60$\pm$0.43
  & 36.17$\pm$0.77
  & 62.61$\pm$0.55
  & 14.82$\pm$0.59 \\

\midrule

\rowcolor{blue!7}
Qwen3.5-9B \emph{(base model)}
  & 55.25
  & 64.86
  & 63.11
  & 44.72 \\

\midrule

Universal Self-Consistency~\citep{chen2023universal}
  & 56.77$\pm$0.89
  & 66.84$\pm$1.28
  & 59.86$\pm$1.96
  & 41.37$\pm$6.16 \\

Policy LLM-Judge~\citep{yuan2024self}
  & \underline{57.33}$\pm$0.50
  & \underline{67.35}$\pm$1.77
  & 60.44$\pm$2.20
  & 46.73$\pm$5.21 \\

Video-Zero~\citep{zhang2026video}
  & 55.06$\pm$0.13
  & 64.80$\pm$0.21
  & 62.91$\pm$0.92
  & 43.30$\pm$0.83 \\

Visplay~\citep{he2025visplay}
  & 55.34$\pm$0.99
  & 65.00$\pm$0.56
  & \underline{64.96}$\pm$0.66
  & 42.55$\pm$1.66 \\

Rubrics as Rewards~\citep{gunjal2025rubrics}
  & 55.60$\pm$0.03
  & 65.03$\pm$0.17
  & 64.72$\pm$1.58
  & \underline{56.45}$\pm$0.95 \\

RLER~\citep{shao2025dr}
  & 56.31$\pm$0.63
  & 65.19$\pm$0.33
  & 64.52$\pm$0.04
  & 56.12$\pm$4.26 \\

RLAC~\citep{wu2025rlac}
  & 56.75$\pm$0.66
  & 67.15$\pm$0.03
  & 61.96$\pm$0.40
  & 55.27$\pm$0.71 \\

\specialrule{1pt}{2pt}{2pt}

\textbf{\method (ours)}
  & \textbf{58.48$\pm$1.59}
  & \textbf{70.93$\pm$1.13}
  & \textbf{69.46$\pm$2.21}
  & \textbf{75.38$\pm$3.32} \\

\bottomrule

\end{tabular}
}

\label{tab:results_with_std_dev}

\end{table}

\subsection{Full PrefEval Results}

\begin{table}[htbp]

\small

\centering

\caption{Per-form PrefEval results. Accuracy is judge
accuracy on the test split. We report the mean with standard deviation $\pm$ over three independent training runs. \textbf{Avg.} is the
macro-average over the three forms and is the value reported in
Table~\ref{tab:results_with_std_dev} and Table~\ref{tab:main_results}. Bold denotes the best result, second best results are underlined.}


\renewcommand{\arraystretch}{1.1}

\setlength{\tabcolsep}{5pt}

\resizebox{\linewidth}{!}{%
\begin{tabular}{lcccc}

\toprule

\textbf{Method}

& \textbf{Explicit}

& \textbf{Choice}

& \textbf{Persona}

& \textbf{Avg.} \\

\midrule

\rowcolor{blue!7}

Qwen3.5-9B \emph{(base model)}

& 55.28

& 39.20

& 39.70

& 44.72 \\

\cdashline{1-5}[0.5pt/1.5pt]

Universal Self-Consistency~\citep{chen2023universal}

& 62.31$\pm$9.95

& 33.92$\pm$1.78

& 27.89$\pm$6.75

& 41.37$\pm$6.16 \\

Policy LLM-Judge~\citep{yuan2024self}

& 57.03$\pm$5.33

& 39.95$\pm$4.62

& 43.22$\pm$5.69

& 46.73$\pm$5.21 \\

Video-Zero~\citep{zhang2026video}
& 51.26$\pm$2.84
& 40.70$\pm$1.42
& 37.94$\pm$1.07
& 43.30$\pm$0.83 \\

Visplay~\citep{he2025visplay}
& 50.50$\pm$3.20
& 41.71$\pm$1.42
& 35.43$\pm$3.20
& 42.55$\pm$1.66 \\

Rubrics as Rewards~\citep{gunjal2025rubrics}
& \textbf{65.33}$\pm$0.71
& 51.26$\pm$0.13
& \underline{52.74}$\pm$3.55
& \underline{56.45}$\pm$0.95 \\

RLER (Evolving Rubrics)~\citep{shao2025dr}
& \underline{63.32}$\pm$7.82
& \underline{53.27}$\pm$0.02
& 51.76$\pm$4.97
& 56.12$\pm$4.26 \\

RLAC (Adversarial Critic)~\citep{wu2025rlac}

  & 60.05$\pm$0.35
  & 53.26$\pm$2.13
  & 52.51$\pm$0.35
  & 55.27$\pm$0.71 \\

\specialrule{1pt}{2pt}{2pt}

\textbf{\method (ours)}

& 61.56$\pm$10.30

& \textbf{80.15$\pm$3.90}

& \textbf{84.42$\pm$3.55}

& \textbf{75.38$\pm$3.32} \\








\bottomrule

\end{tabular}
}

\label{tab:prefeval_full}

\end{table}

\subsection{Test-Time Training}
\label{app:ttt}
We further evaluate \method under the standard \emph{test-time training} (TTT) setting, where the model adapts to the unlabeled test distribution. Following the protocol of TTRL~\citep{zuo2026ttrl}, we initialize from the base model and treat the test split as unlabeled adaptation data, with gold annotations withheld throughout adaptation and used only for final evaluation. We compare \method against TTRL under the same base-model initialization, test inputs, GRPO optimization, and adaptation-cycle budget (scaled according to test-set size; three passes for IntentQA, UR-FUNNY, and PrefEval, and two for the larger MOSEI test set). \method improves over the base model on all four benchmarks, whereas TTRL's majority-vote pseudo-labels may be unreliable here. TTRL helps in some settings but falls below the base model in others, and collapses on free-form PrefEval, more severely with the larger rollout group. This is consistent with the known failure mode of consensus rewards when the majority answer is wrong or when free-form answers rarely coincide, so that voting reinforces a degenerate mode; \method may avoid this by ignoring rollout frequency.

\begin{table}[h]
\small
\centering
\caption{Test-time training performance. All methods are initialized from the same Qwen3.5-9B base model. Following TTRL~\citep{zuo2026ttrl}, we scale test-time adaptation with test-set size: three passes for IntentQA, UR-FUNNY, and PrefEval, and two for the larger MOSEI test set, using the same dataset-specific budget across methods. $G$ denotes rollout group size; we use $G{=}4$ for \method and report TTRL with $G{=}4$ and an expanded-budget $G{=}16$ setting. $^{\dagger}$PrefEval is the macro-average over its three preference tasks. Bold denotes the best result, second best results are underlined.}
\renewcommand{\arraystretch}{1.05}
\setlength{\tabcolsep}{10pt}
\resizebox{\linewidth}{!}{%
\begin{tabular}{lcccc}
\toprule
\textbf{Method}
& \textbf{IntentQA} ($\uparrow$)
& \textbf{MOSEI (S)} ($\uparrow$)
& \textbf{UR-FUNNY} ($\uparrow$)
& \textbf{PrefEval}$^{\dagger}$ ($\uparrow$) \\
\midrule

\rowcolor{blue!7}
Qwen3.5-9B \emph{(base)}
& 55.25
& 64.86
& 63.11
& 44.72 \\

\cdashline{1-5}[0.5pt/1.5pt]

TTRL ($G{=}4$)~\citep{zuo2026ttrl} 
& 56.00
& 63.47
& \underline{65.81}
& \underline{36.52} \\

TTRL ($G{=}16$)~\citep{zuo2026ttrl} 
& 53.84
& \underline{68.81}
& 60.13
& 18.26 \\

\specialrule{1pt}{2pt}{2pt}

\textbf{\method ($G{=}4$) (ours)}
& \textbf{58.34}
& \textbf{70.65}
& \textbf{70.33}
& \textbf{73.70} \\

\bottomrule
\end{tabular}%
}
\label{tab:ttt_results}
\end{table}

\begin{table}[htbp]
\centering
\caption{Test-time training performance on the three PrefEval task settings. All methods start from the same Qwen3.5-9B base model. We report the final score after three passes for \method and TTRL with $G{=}4$, and the third-pass score for TTRL with $G{=}16$. Avg. is the macro-average across Explicit, Choice, and Persona. Bold denotes the best result, second best results are underlined.}
\renewcommand{\arraystretch}{1.1}
\setlength{\tabcolsep}{25pt}

\resizebox{\linewidth}{!}{%
\begin{tabular}{lcccc}

\toprule

\textbf{Method}
& \textbf{Explicit}
& \textbf{Choice}
& \textbf{Persona}
& \textbf{Avg.} \\

\midrule

\rowcolor{blue!7}
Qwen3.5-9B \emph{(base model)}
& 55.28
& 39.20
& 39.70
& 44.72 \\

\cdashline{1-5}[0.5pt/1.5pt]

TTRL ($G{=}4$)
& \textbf{52.26}
& 34.17
& 23.12
& 36.52 \\

TTRL ($G{=}16$)
& 21.61
& 19.60
& 13.57
& 18.26 \\

\specialrule{1pt}{2pt}{2pt}


\textbf{\method ($G{=}4$) (ours)}
& \underline{49.75}
& \textbf{81.41}
& \textbf{89.95}
& \textbf{73.70} \\

\bottomrule

\end{tabular}
}

\label{tab:prefeval_ttt}

\end{table}


\clearpage
\section{RLCE Implementation Details}
\label{app:rlcv-implementation}

This section provides full implementation details of \method,
including the exact training configuration, reward specification, and prompts
used for comparative evidence.

\subsection{Implementation Details}
\label{app:verification_details}

We use Qwen3.5-9B for both the trainable policy $\pi_\theta$ and the frozen
verifier $\pi_f$. The policy $\pi_\theta$ is fully fine-tuned, while $\pi_f$ remains fixed
throughout training and is used for free-text canonicalization, evidence-test
construction, and evidence test validation. No stronger external model or
ground-truth annotation is used for reward construction. 


For free-text outputs, the answers are
canonicalized before verification. Candidate pairs are judged for semantic equivalence in a single greedy, schema-constrained call to $\pi_f$. Closed-label outputs are mapped directly to their
corresponding answer options.

Evidence-test construction uses temperature $1.0$, top-$p=0.95$, and
top-$k=20$, while canonicalization and evidence validation use greedy
decoding. We construct one initial evidence test per direction for IntentQA
and PrefEval, three for MOSEI-S, and six for UR-FUNNY. Candidate answers are
removed from the validator input before evidence test validation.



\subsection{Reward and GRPO Optimization}
\label{app:rlcv-grpo}

For rollout $a_g$, the implemented reward is
\[
r_g
=
0.8\,r_g^{\mathrm{ver}}
+
0.2\,r_g^{\mathrm{fmt}},
\]
where
\[
r_g^{\mathrm{ver}}
=
\mathbf{1}
\left[
\operatorname{canon}(a_g)=\hat a
\right]
\]
indicates that the rollout canonicalizes to the verified answer $\hat a$, and
\[
r_g^{\mathrm{fmt}}
=
\mathbf{1}
\left[
a_g
\text{ ends in the required }
\backslash\texttt{boxed\{\}}
\text{ answer format}
\right].
\]
The format reward is purely structural and contains no information about answer correctness; it only encourages compliance with RLCE's training-time response schema. At evaluation, task answers are extracted from each method's corresponding response format before applying the common benchmark metric. When verification abstains, or when a free-text rollout
group contains only one canonical answer, $r_g^{\mathrm{ver}}=0$ for all
rollouts.

For a rollout group of size $G$, let
\[
\mu_G
=
\frac{1}{G}\sum_{g=1}^{G}r_g,
\qquad
\sigma_G
=
\sqrt{
\frac{1}{G}
\sum_{g=1}^{G}(r_g-\mu_G)^2
}.
\]
The group-relative advantage is
\[
A_g
=
\begin{cases}
\dfrac{r_g-\mu_G}{\sigma_G},
& \sigma_G>0,\\[6pt]
0,
& \sigma_G=0.
\end{cases}
\]
Thus, rollout groups with no reward variation contribute no relative
policy-gradient signal.

We optimize the policy using Group Relative Policy Optimization
(GRPO)~\citep{shao2024deepseekmath}. Let
$\pi_{\theta_{\mathrm{old}}}$ denote the rollout policy and define the
per-token importance ratio
\[
\rho_{g,t}(\theta)
=
\frac{
\pi_\theta(a_{g,t}\mid c,q,a_{g,<t})
}{
\pi_{\theta_{\mathrm{old}}}(a_{g,t}\mid c,q,a_{g,<t})
}.
\]
The clipped GRPO objective is
\begin{align}
\mathcal J_{\mathrm{GRPO}}(\theta)
=
\frac{1}{G}
\sum_{g=1}^{G}
\frac{1}{|a_g|}
\sum_{t=1}^{|a_g|}
\Big[
&
\min\!\Big(
\rho_{g,t}(\theta)A_g,\,
\operatorname{clip}
\big(
\rho_{g,t}(\theta),
1-\epsilon,
1+\epsilon
\big)A_g
\Big)
\nonumber\\[-2pt]
&
-\beta D_{g,t}^{\mathrm{KL}}
\Big].
\end{align}
We use $\beta=0.01$ and VERL's \texttt{low\_var\_kl} estimator, with the
frozen base model as the reference policy. The KL term is applied directly
in the policy loss rather than incorporated into the rollout reward.

We train using VERL with $G=4$ rollouts per prompt, full-parameter
fine-tuning, AdamW, a learning rate of $1\times10^{-6}$ with a constant
schedule and no warm-up, one PPO epoch per batch, and no entropy bonus.
GRPO clipping uses the VERL default and is not overridden. Rollouts use
temperature $1.0$, top-$p=0.95$, top-$k=20$, presence penalty $1.5$,
repetition penalty $1.0$, and a maximum response length of 512 tokens.

The number of sampled training examples per cycle is 96 for IntentQA, 96 for
UR-FUNNY, 320 for MOSEI-S, and 48 for PrefEval, with PPO mini-batch sizes of
16, 16, 32, and 16, respectively. The maximum numbers of training cycles are
40, 25, 45, and 20. Maximum prompt length is 4096 tokens for IntentQA,
UR-FUNNY, and MOSEI-S and 8192 tokens for PrefEval. We utilize 8 frames per clip at 1 FPS.



\subsection{\method Prompts}
\label{app:rlcv-prompts}

\textbf{System prompts for evidence-test construction.} No ground-truth annotations are used in the prompts for \method or any baseline. Evidence-test construction uses a fixed prompt that takes the input sample (e.g., the video clip and transcript), question, and candidate answers as input and asks the model to produce checkable comparative evidence tests (Figure~\ref{fig:evidence-test-author-prompts}). The same construction prompt is used across examples, with only the example-specific inputs substituted. Such fixed prompting is standard in instruction-based learning and model-based evaluation~\citep{wei2021finetuned,mishra2022cross,bai2022constitutional,liu2023geval,yuan2024self,gunjal2025rubrics}. Controlled ablations in Section~\ref{sec:ablations} examine the method's individual design choices while holding the construction prompt, validation model, and remaining training configuration fixed.

\textbf{Evidence testing prompts.} Evidence tests are evaluated against the observed sample using a simple validation prompt (Fig.~\ref{fig:evidence-test-validator-prompt}). Given the sample and a set of evidence tests, the validator judges each test as \texttt{yes}, \texttt{no}, or \texttt{unknown} according to whether it is supported by the available input. The same general validation procedure is used across tasks; Fig.~\ref{fig:evidence-test-validator-prompt} shows a representative prompt.

\paragraph{Solver and canonicalization prompts.}
The trainable policy uses the solver prompt in Fig.~\ref{fig:solver-system-prompt}. For free-text outputs, all remaining candidate-pair equivalence decisions after deterministic duplicate merging are obtained in a single greedy call to the frozen verifier using Fig.~\ref{fig:canonicalization-prompt}.

\begin{figure}[htbp]
\centering
\begin{promptbox}[breakable=false]{{Evidence-Test Construction Prompt}}

\{Input sample (e.g., video clip and transcript):\}

Question: \{question\}

Canonical candidate answers for this pair:
G1: "..."
G2: "..."

Write only the requested comparative evidence tests. Each test must describe a concrete fact that is more expected under \texttt{for\_group} than \texttt{against\_group}, must be checkable from the provided input, and must not restate the question or either candidate answer. The rationale should explain the direction, but its wording is diagnostic only.

Return only a JSON array of objects with exactly:
\texttt{pair\_id}, \texttt{for\_group}, \texttt{against\_group},
\texttt{evidence\_test}, and \texttt{rationale}.
\end{promptbox}
\caption{Example prompt used to construct evidence tests.}
\label{fig:evidence-test-author-prompts}
\end{figure}

\begin{figure}[htbp]
\centering
\begin{promptbox}[breakable=false]{{Evidence-Test Validation Prompt}}
You are checking observable statements about a video clip.

Sample:  
\{sample\}

Statements:  
\{evidence tests\}

For each statement, use only the clip and transcript, if available. Reply \texttt{yes} if the available evidence supports the statement, \texttt{no} if it contradicts it, and \texttt{unknown} if the evidence is insufficient. Return one numbered line per statement, for example:

1: yes\\
2: no\\
3: unknown
\end{promptbox}
\caption{Example of the prompt used to validate evidence tests against the observed sample.}
\label{fig:evidence-test-validator-prompt}
\end{figure}

\begin{figure}[htbp]
    \centering
    \begin{promptbox}[breakable=false]{{Solver System Prompt}}
You are the Solver. You are given frames sampled from a video clip and the transcript of what is said --- either may be absent --- together with one question about it. No audio track is provided, even where the question mentions one.

Your task is to answer the question.

Rules:
1. Base your judgment only on the frames and the transcript.
2. If the question lists answer options, answer with exactly one of them, written as given. Otherwise answer in the form the question asks for.

Reason briefly, then end your response with the answer and nothing after it:
\texttt{\textbackslash boxed\{<your answer>\}}
    \end{promptbox}
    \caption{System prompt used by the \method trainable policy during training and evaluation.}
    \label{fig:solver-system-prompt}
\end{figure}






\begin{figure}[htbp]
\centering
\begin{promptbox}[breakable=false]{{Canonicalization Prompt}}
Question:  
\{question\}

Candidate answers:  
\{candidate answers\}

For each listed pair, determine whether the two candidates are semantically equivalent answers to the question. Treat ordinary paraphrases and wording differences as equivalent.

Return only the equivalence decision for each pair in the specified JSON format.
\end{promptbox}
\caption{Example of the prompt used to canonicalize free-text rollout answers.}
\label{fig:canonicalization-prompt}
\end{figure}

\clearpage
\section{Baseline Implementation}
\label{apx:baselines}

In this section, we detail the implementation for each baseline. We adapted five current methods for our baselines: Video-Zero~\citep{zhang2026video}, VisPlay~\citep{he2025visplay}, Evolving Rubrics (RLER)~\citep{shao2025dr}, Static rubrics (Rubrics as Rewards, RaR)~\citep{gunjal2025rubrics}, RLAC Adversarial Critic~\cite{wu2025rlac}, Universal Self Consistency (USC)~\citep{chen2023universal}, Policy LLM-Judge (as adapted from Self Rewarding Language Models~\citep{yuan2024self}). These methods were chosen to be comparable since they share this notion of learning with unlabeled training data. Appendix~\ref{app:additional_results} detail the tables for the overall performance of the baselines reported as the mean and its corresponding standard deviation over three runs.



\subsection{Universal Self-Consistency}

We implement Universal Self-Consistency (USC) following
\citet{chen2023universal}, which determines the preferred answer from agreement
among the policy's own sampled responses rather than by evaluating those
responses against external supervision. USC operates on the same rollout groups
used for RL training and receives no ground-truth annotations.

For closed-set tasks, where rollout answers can be aggregated directly, we
compute the vote count over all parseable predictions and select the unique
most frequent answer. Rollouts matching the selected answer receive reward 1
and all others receive reward 0. If the highest vote count is tied, the group
receives no reward, since the rollouts do not determine a unique consensus
answer. No additional model generation is required in this setting.

For free-form tasks, exact-match voting is insufficient because responses with
the same underlying answer may differ in surface form. Following the USC
procedure of \citet{chen2023universal}, we therefore provide the question and
all parseable responses in the rollout group to the frozen LLM and ask it to
select the response most consistent with the majority of the candidates. The
selector receives only the question and candidate responses; the underlying
multimodal context, preference evidence, answer options, and ground-truth
answer are not included. We use one greedy selection per rollout group. All
rollouts whose response exactly matches the selected response receive reward 1,
while the remaining rollouts receive reward 0. Thus, USC derives its training signal solely from agreement among the
policy's current rollouts. We provide the prompt that we utilized in Figure~\ref{fig:usc-prompt}.


USC uses the same RL training setup as \method, including the model
initialization, rollout and data budgets, optimization and sampling
hyperparameters, KL regularization, and auxiliary format reward. Only the task
reward differs. USC derives it from agreement among the policy's sampled
responses.

\begin{figure}[htbp]
    \centering
    \begin{promptbox}[breakable=false]{{Universal Self-Consistency Prompt}}
I have generated the following \{k\} responses to a question:

Question: \{question\}

\{numbered\}

Select the response that is most consistent with the majority of the responses. Reply with exactly the number of the selected response and nothing else.
    \end{promptbox}
    \caption{Universal Self-Consistency prompt used for free-form tasks. Following \citet{chen2023universal}, the selector receives the question and sampled responses and selects the response most consistent with the majority.}
    \label{fig:usc-prompt}
\end{figure}

\subsection{Policy LLM-Judge}

We implement a pointwise Policy LLM-Judge baseline adapted from the
self-rewarding mechanism of \citet{yuan2024self}. In Self-Rewarding Language
Models, the policy itself acts as an LLM-as-a-Judge: candidate responses are
evaluated individually and assigned scalar rewards, which are subsequently used
to construct preference supervision. For our case, we utilize a frozen policy of the base Qwen3.5-9B model. Evaluating individual responses
pointwise is also a standard LLM-as-a-Judge formulation
\citep{zheng2023judging}. We retain this direct self-judging setup while
adapting the judgment to our task setting and binary GRPO reward.

For closed-set tasks, each candidate answer is judged independently. The
frozen policy receives the question, the available input evidence, and a single
proposed answer, and is asked whether that answer is correct. The judge is not
shown the remaining answer options. A parsed ``Yes'' assigns reward 1 to
rollouts producing that answer; all other outcomes receive reward 0. Each distinct option is judged once per example
and the resulting verdict is shared by all rollouts predicting that option.
This pointwise construction follows the single-response LLM-as-a-Judge setup. 


For free-form tasks, we apply the same pointwise judgment to each distinct
rollout response. The frozen policy receives the question, available input
evidence, and proposed response and judges whether the response is correct.
Duplicate responses are deduplicated before evaluation so that identical
answers receive identical rewards. A parsed ``Yes'' receives reward 1;
otherwise the reward is 0.

For both closed-set and free-form tasks, the judge is a frozen copy of the base
model (Qwen3.5-9B), which is the same as the validator in \method, and ground-truth training annotations are not provided during reward
construction. This keeps the comparison focused on the source of the training
signal: direct policy judgment for Policy LLM-Judge, agreement among policy
rollouts for USC, and comparative evidence testing for \method.

Policy LLM-Judge uses the same RL training setup as \method, including the model
initialization, rollout and data budgets, optimization and sampling
hyperparameters, KL regularization, and auxiliary format reward. Only the task
reward differs. Policy LLM-Judge derives it from pointwise judgments by the
frozen base policy.

\begin{figure}[htbp]
    \centering
    \begin{promptbox}[breakable=false]{{Policy LLM-Judge Prompt}}
You are checking whether an answer to a question about a video clip is correct.

Transcript:
\{transcript\}

Question: \{question\}

Proposed answer: \{answer\}

Based on the video frames and the transcript, if any, is the proposed answer correct? Reply with exactly one word: yes or no.
    \end{promptbox}
    \caption{Pointwise Policy LLM-Judge prompt used to evaluate each candidate answer independently. The judge receives the question, available input evidence, and a single proposed answer, and returns a binary correctness judgment.}
    \label{fig:policy-llm-judge-prompt}
\end{figure}

\subsection{Evolving Rubrics (RLER) \& Static Rubrics as Rewards (RaR)}
\label{apx:evolving_rubrics_implementation}

This section details how we implement the rubrics-based baselines. We follow the convention and code implemented in DR Tulu~\citep{shao2025dr}. In particular, we use the same backbone model (e.g., Qwen 3.5 9B) for three roles: a policy model to be trained; a (frozen) rubric generator model; and a (frozen) judge model to score the rollouts against the rubrics. The policy model trains with GRPO with the rewards being derived from a buffer of generated (persistent and active) rubrics. While the original task is for deep research, we adapted the method, of evolving rubrics, for social understanding tasks. 

\textbf{Rubric buffer.} Each training clip has its own buffer with three parts: a \emph{persistent} set of 3--5 seed rubrics, written once before training by the generator from the clip and its transcript; an \emph{active} set that starts empty and is generated online during training (capped at 5 sets); and an archive of pruned rubrics. Every rubric carries a weight, $+1$ for a positive rubric (a positive trait in the response) and $-1$ for a negative one (a negative trait in the response). In Evolving Rubrics, each visit to a clip runs the full loop of Algorithm~\ref{alg:evolving-rubrics}: the policy emits $G=8$ rollouts, the rubric generator proposes new candidate rubrics, the judge scores every rubric against every rollout, the rewards feed the GRPO step, and a prune removes active rubrics whose scores have zero standard deviation over the rollouts. We do not prune the persistent rubrics as outlined in Dr Tulu. In the Static Rubrics as Rewards baseline, the active set stays empty: the rubric generator is called only in the seed pass, and the judge scores the persistent set throughout the training. Gold labels were not used anywhere in either pipeline.

\begin{algorithm}[htbp]
\caption{Evolving Rubrics (Static Rubrics skips lines \ref{alg:er:gen} and \ref{alg:er:prune})}
\label{alg:evolving-rubrics}
\begin{algorithmic}[1]
\Require Backbone $\theta_{0}$; train split $\mathcal{C}$; validation split; epoch cap $E=4$
\Ensure Trained Policy $\pi$
\State $\pi \gets \theta_{0}$; \quad Generator, Judge $\gets \theta_{0}$ \Comment{Generator and Judge frozen}
\For{$c \in \mathcal{C}$}
    \State $P_c \gets \textsc{Seed}(c)$; \quad $A_c \gets \emptyset$ \Comment{3--5 persistent seed rubrics}
\EndFor
\For{$e = 1, \dots, E$}
    \For{each batch of 32 clips; each clip $c$ in the batch}
        \State $y_{1}, \dots, y_{8} \gets$ Policy rollouts on $c$
        \State $N_c \gets \textsc{Generate}(c,\, y_{1..8},\, P_c \cup A_c)$ \Comment{0--5 new rubrics, weighted $\pm 1$} \label{alg:er:gen}
        \State $J(r, y_i) \gets$ Judge score for every $r \in P_c \cup A_c \cup N_c$, every $y_i$
        \State $r_i \gets 0.8\, S_{\mathrm{rubric}}(y_i) + 0.2\, r_{\mathrm{format}}(y_i)$
        \State $A_c \gets \textsc{Prune}(A_c \cup N_c)$ \Comment{drop zero-variance rubrics, keep top 5 by score std} \label{alg:er:prune}
        \State $\pi \gets \textsc{GRPO}(\pi, \text{batch rewards})$ \Comment{one optimizer step per batch}
    \EndFor
    \State Validate every 12 steps; stop after 6 validations without improvement
\EndFor
\State \Return best validation checkpoint of $\pi$
\end{algorithmic}
\end{algorithm}

\textbf{Training.} We run one pipeline per task on the task's full training split and evaluate the checkpoint with the best validation score under the common evaluation procedure (Appendix~\ref{apx:evaluation_implementation}). The answer space is the task's own label set; IntentQA and PrefEval can contain free text.

The transcript appears in the policy prompt, the seed pass, and the per-visit generator prompt. For IntentQA the prompts omit the transcript section and the seed generator grounds in the clip's frames instead; its question is per-clip, joined from the training split. The closed-label tasks share one task question per dataset in the training split (e.g., \texttt{What is the sentiment of the speaker in this video?} for MOSEI-Sentiment). The PrefEval tasks have no video and the transcript in these tasks are instead replaced by the user-assistant conversation.

\begin{figure}
    \centering
    \begin{promptbox}[breakable=false]{Policy Template}
        Watch the video and answer the question below.\\[4pt]
        Question: \{question\}\\[4pt]
        Transcript\\
        \{transcript\}\\[4pt]
        Give your final answer inside <answer> </answer> tags. Answer with exactly one of: \{labels\}. Then, in <justification> </justification> tags, give 1 to 3 sentences of evidence from the clip that support your answer.\\[4pt]
        Your output format must be:\\
        <answer>...</answer>\\
        <justification>...</justification>
    \end{promptbox}
    \label{fig:prompt-policy-er}
\end{figure}

Following the prompt used in DR Tulu, the justification is the text the rubrics score on. For open ended tasks such as IntentQA, the label sentence becomes \texttt{Answer in one short sentence.}

\textbf{Seed rubrics.} Before training, a seed pass runs the generator once per clip with the prompt below to write that clip's persistent rubrics. Static Rubrics uses this set, unchanged, for the whole run.

\begin{figure}
    \centering
    \begin{promptbox}[breakable=false]{Seed Generator Template}
        You are an expert evaluator writing a small set of initial rubrics to judge how well a short response answers a question about a short video clip.\\[4pt]
        Question: \{question\}\\
        The response gives one of these labels: \{labels\}, plus a 1-3 sentence evidence justification.\\[4pt]
        You are given the clip's transcript below as grounded, privileged evidence. Write rubrics that a strong 1-3 sentence justification of the answer should satisfy -- specific, transcript-grounded criteria about the evidence a good answer cites, not generic style, length, or verbosity.\\[4pt]
        \#\# Transcript\\
        \{transcript\}\\[4pt]
        \#\# Instructions\\
        - Write between 3 and 5 rubrics (no more than 5).\\
        - Each rubric states what an excellent justification should contain.\\
        - Ground each rubric in a concrete cue from the transcript where possible.\\
        - Each rubric has a detailed description and a short abstract title.\\[4pt]
        \#\# Output Format\\
        \{"positive\_rubrics": [\{"description": "<what an excellent justification contains>", "title": "<short label>"\}]\}
    \end{promptbox}
    \label{fig:prompt-seed-er}
\end{figure}

\textbf{Per-visit rubrics (Evolving only).} On every sample visit, the generator reads the clip (if applicable), the transcript, the 8 rollouts, the current rubric buffer, and proposes up to 5 new positive or negative rubrics. An empty proposal is valid when the pool already covers everything. A rubric gets weight $+1$ if the generator lists it as positive and $-1$ if it lists it as negative. We include two rules in the prompt. First, a negative rubric must describe a failure. The judge is never told whether a rubric is positive or negative. If a negative rubric is written as ``the justification should not merely restate the label'', a good response satisfies it, gets a high score, and is then penalized by the $-1$ weight. Second, the generator must not write a positive and a negative version of the same criterion, because the pair counts the same behaviour twice in the reward sum.

\begin{figure}
    \centering
    \begin{promptbox}[breakable=false]{Rubric Generator Template}
        You are an expert evaluator generating adaptive rubrics to judge how well a short response answers a question about a short video clip. Each response is a label plus a 1-3 sentence evidence justification.\\[4pt]
        Question: \{question\}\\
        The response gives one of these labels: \{labels\}.\\[4pt]
        \#\# Task\\
        Identify the most discriminative criteria that separate high-quality justifications from low-quality ones -- criteria that distinguish the actual responses below, not generic checks that apply to all of them.\\[4pt]
        \#\# Categories\\
        - **Positive rubrics**: excellence indicators a strong justification shows (e.g. cites a concrete spoken or visual cue for its label).\\
        - **Negative rubrics**: clear failure modes that degrade a justification (e.g. asserts a label with no evidence, or farms length and confident-sounding phrasing without citing a real cue).\\[4pt]
        \#\# Guidelines\\
        - Write 1-5 rubrics total; fewer high-quality rubrics beat many generic ones. Return empty lists if the existing rubrics already cover everything.\\
        - Do not duplicate an existing rubric in meaning or scope.\\
        - Ground rubrics in the transcript and the observed differences among the responses.\\[4pt]
        \#\#\# Avoid mirror rubrics\\
        Never write a positive and a negative version of the same criterion; keep only the direction that better separates the responses below. The existing rubrics are listed with their weight: +1.0 is a positive rubric, -1.0 is a negative one, so check both directions before adding a rubric.\\
        - Bad: "Cites a concrete spoken cue for the chosen label" together with "Fails to cite a concrete spoken cue".\\
        - Good: whichever one of the two is more discriminative, on its own.\\[4pt]
        \#\#\# Conservative negative rubrics\\
        A response is penalized if it exhibits ANY negative rubric behavior, so a negative rubric must name that behavior. Identify clear failure modes, not the absence of excellence: focus on active mistakes rather than missing features. Never phrase a negative rubric as a prohibition -- a good response obeys a prohibition and would be penalized for it.\\
        - Bad: "The justification should not merely restate the label."\\
        - Good: "Restates the chosen label as its own justification without pointing to any cue in the clip."\\[4pt]
        \#\# Transcript\\
        \{transcript\}\\[4pt]
        \#\# Responses\\
        Response 1: \{rollout 1\} \dots\ Response 8: \{rollout 8\}\\[4pt]
        \#\# Existing rubrics\\
        \{current pool as a JSON array of \{"description", "weight"\} objects\}\\[4pt]
        \#\# Output Format\\
        \{"positive\_rubrics": [\{"description": ..., "title": ...\}], "negative\_rubrics": [\{"description": ..., "title": ...\}]\}
    \end{promptbox}
    \label{fig:prompt-pervisit-er}
\end{figure}

\textbf{Judging.} The judge model (frozen Qwen 3.5 9B) is text-only: it will only receive a generic task question, the full policy response, and one rubric per prompt to score on. Its scoring sentence is the prompt implemented in  DR Tulu (how \emph{appropriate} the response is against the criterion). It returns an integer in $\{0, 1, 2\}$, divided by 2 to $J(r, y) \in \{0, 0.5, 1\}$; a reply with no parseable score, or an empty response, scores 0.

\begin{figure}
    \centering
    \begin{promptbox}[breakable=false]{Judge Template}
        You will be given a question someone was asked (in <question></question> tags) and the corresponding response (in <response></response> tags). You will then be given a specific criterion to evaluate (in <criterion></criterion> tags).\\
        Return a score on a scale of 0 to 2 indicating how appropriate the response is based on the given criterion. Judge only the specified criterion, not any other qualities of the response. Output JSON in the format: \{"score": x\}.\\[4pt]
        <question>\{task question\}</question>\\
        <response>\{full policy response\}</response>\\
        <criterion>\{rubric\}</criterion>
    \end{promptbox}
    \label{fig:prompt-judge-er}
\end{figure}

The \{task question\} slot in the judge template is the dataset's question plus a response-shape sentence (\texttt{Answer with exactly one of: \{labels\}, and support the answer with 1 to 3 sentences of evidence from the clip.}), identical for every clip of a closed-label task. A rollout's rubric score follows the original work's Eq.~1: with rubric weights $w_k \in \{+1, -1\}$,
\begin{equation*}
S_{\mathrm{rubric}}(y) \;=\; \frac{\sum_{k} w_k \, J(r_k, y)}{\max\bigl(\sum_{k :\, w_k > 0} w_k,\; 1\bigr)},
\end{equation*}
where the sum runs over the clip's persistent, active and candidate rubrics (active and candidate rubrics only applicable to evolving rubrics). The denominator counts only positive weights; $S_{\mathrm{rubric}}$ is not clamped and can fall below 0 when penalties dominate.

\textbf{Rewards.} Each rollout is scored
\begin{equation*}
r \;=\; 0.8\, S_{\mathrm{rubric}} \;+\; 0.2\, r_{\mathrm{format}},
\end{equation*}
where $r_{\mathrm{format}}$ is 1 when the response carries an \texttt{<answer>} tag parsing to exactly one label of the task's space and a non-empty justification, and 0 otherwise; for the free-text task the check is structural only (non-empty answer and justification). The original work blends four terms, rubric, citation, format, and search, and this task has no analog for citation or search, so their weights do not carry over; the 0.8/0.2 split is our own choice, and only the ratio matters because the GRPO advantage is normalized by the group standard deviation.

\textbf{Pruning.} At the end of each visit, every active rubric and candidate is scored by the population standard deviation of its Judge scores across the 8 rollouts. Rubrics with zero standard deviation, those that separated no rollout from another, are dropped, and of the rest the top 5 by standard deviation are kept as the new active set. Persistent rubrics are exempt, and culled rubrics are archived. This mechanism retains discriminative criteria while dropping stale criteria as stated in the original work.

\begin{table}[tbp]
\centering
    \caption{Optimization settings for Evolving and Static Rubrics. We follow the original work's settings wherever applicable.}
    \label{tab:evolving-rubrics-optimization-settings}
    \begin{tabular}{lc}
    \toprule
    \textbf{Setting} & \textbf{Policy} \\
    \midrule
    Optimization algorithm            & GRPO               \\
    Learning rate                     & $5 \times 10^{-7}$ \\
    KL coefficient                    & 0.001              \\
    Epoch cap                         & 4                  \\
    Unique clips per rollout batch    & 32                 \\
    Rollouts per clip for GRPO        & 8                  \\
    Optimizer steps per rollout batch & 1                  \\
    Validation interval (steps)       & 12                 \\
    Early-stopping patience           & 6 val runs      \\
    Checkpoint used for test          & best val run    \\
    Rollout temperature               & 1.0                \\
    Generator / Judge temperature     & 0.0                \\
    Max response length               & 1024               \\
    Training precision                & bfloat16           \\
    Rollout engine                    & vLLM               \\
    \bottomrule
    \end{tabular}
\end{table}

\paragraph{Limitations of rubric rewards.} In our experiments, we observed several limitations of rubric rewards, which motivate the use of evidence in \method. \label{apx:limitation_rubrics}

\begin{figure}[h]
    \centering
    \includegraphics[width=\linewidth]{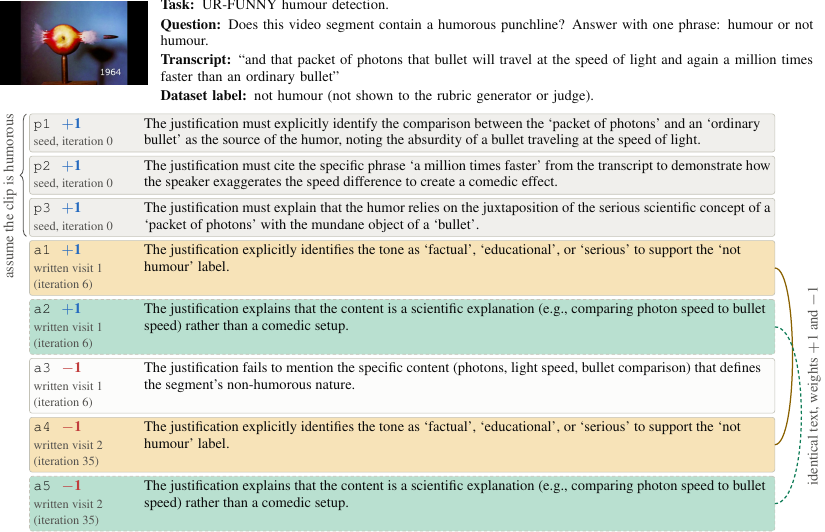} 
    \setlength{\emergencystretch}{8em}
    \caption{An example of \textbf{conflicting criteria} in RLER. UR-FUNNY clip \texttt{train\_00088\_00122} (label: \texttt{not humour}). At visit 2 the generator wrote \texttt{a4} and \texttt{a5}: word-for-word copying the two active criteria \texttt{a1} and \texttt{a2} with the sign reversed. \texttt{a1} and \texttt{a2} both reward arguing for ``not humour'' and agreeing with the label; even worse, the three seeded persistent rubrics (\texttt{p1}, \texttt{p2}, \texttt{p3}) \textbf{incorrectly} assume the clip is humorous. Note that the pruning mechanism for RLER only checks for the standard deviation of past rewards for each criterion; thus, despite this conflict, these criteria may not get pruned. \texttt{a6}, \texttt{a7} and \texttt{a8} (weight $+1$, not shown) reword the three seeds; they scored 0 on every rollout and were culled at visit 2, but they still count in $P$.}

    \label{fig:conflicting_criteria}
\end{figure}

\begin{figure}[t]
    \centering
    \includegraphics[width=\linewidth]{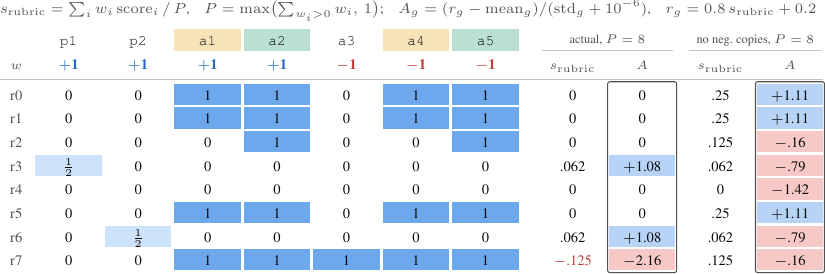} 
    \setlength{\emergencystretch}{8em} 
    \caption{\textbf{Criterion Collapse:} conflicting or near-redundant criteria neutralize relative advantages, inhibiting the effectiveness of GRPO. Following the example in Figure~\ref{fig:conflicting_criteria}, each conflicting pair contributes $w_i\,s_i + w_j\,s_j = 0$ to the numerator; yet, it contributes ($+1$) to $P$. This affects normalization; and in turn, dilutes the relative advantages of individual rollouts. Without the negative (conflicting) copies, the relative advantages (second column) may become more discriminative; thus, carrying a stronger training signal for GRPO.}
    \label{fig:criterion_collapse}
\end{figure}

We observe that the Rubric Generator can occasionally produce imperfect or conflicting criteria in both the persistent and active rubric sets. Such inconsistencies can reduce the usefulness of the resulting GRPO signal. Figure~\ref{fig:conflicting_criteria} shows examples of conflicting criteria, with the corresponding reward behavior illustrated in Figure~\ref{fig:criterion_collapse}. When opposing criteria assign similar rewards to competing responses, the resulting relative advantages can be reduced or neutralized. We refer to this behavior as criterion collapse (Figure~\ref{fig:criterion_collapse}).

One possible explanation is the difference in task structure. RLER was introduced in DR Tulu for deep-research settings, where responses are substantially longer and may support a broader range of useful evaluation criteria. In our setting, outputs are comparatively short, typically consisting of a label and one to three sentences of evidence, which may provide fewer dimensions along which evolving rubrics can differentiate candidate responses. This may make criterion generation less informative or more susceptible to overlap and conflict. Static rubrics avoid changes in the active criterion set during training, but can still inherit limitations from the initially generated criteria. These observations motivate \method's use of evidence derived locally from the observed input and the current candidate answers, rather than relying on an expanding set of task-level criteria.

\textbf{Optimization and compute.} The Policy trains with GRPO on verl: group-normalized advantages, the clipped surrogate, and a KL loss against the reference, with DR Tulu's optimizer settings. Each rollout batch holds 32 unique clips with $G = 8$ rollouts each, and we take one optimizer step per batch. verl drops the last partial batch, so one epoch is $\lfloor |\mathcal{C}| / 32 \rfloor$ steps.

Training is capped at 4 epochs. The original work's open-judge ablation, which also uses the policy's base model as generator and judge, runs 1000 steps (about 3.6 epochs); we round this to 4. Every 12 optimizer steps we score the current checkpoint on the full validation split of each task with the same settings as the test-split evaluation. We then keep the checkpoint with the best validation score during training, or stop early once 6 consecutive validations fail to improve on it. This best checkpoint is the one evaluated on test set of each task.

The policy model samples at temperature 1.0; the generator and judge decode greedily; thinking is off for every role. Video is sampled at 1 fps, 1 to 8 frames, resized to $176 \times 320$, to replicate the settings in the evaluation. 


\textbf{Ensuring a fair implementation for rubrics-based methods.} Both rubric baselines run DR Tulu's released implementation with the optimizer settings of the original work (Table~\ref{tab:evolving-rubrics-optimization-settings}) and its open-judge setting: both the rubric generator and the judge are the policy's own base model (Qwen 3.5 9B). Where the original recipe leaves a choice open, we again resolve it in the baseline's favor. Both baselines train on the full training split of each task, with persistent rubrics seeded for every clip before training starts (e.g., 16{,}210 generations on CMU-MOSEI); the 4-epoch cap follows the original work's ablation on visits per clip (about 3.6 epochs, rounded up). Notably, this training budget is substantially larger than that used by \method, which sees fewer than 1.5 effective epochs of the training data. The reported checkpoint is the best one on the full validation split, chosen from a validation every twelve optimizer steps, rather than the final one. In compute, a rubric run is the most expensive baseline per optimizer step because every rollout is judged against every rubric in the buffer. Static Rubrics shares the code and settings of RLER and only omits the generated active rubrics.

\clearpage
\subsection{Visplay}
This section details how we implement the VisPlay~\citep{he2025visplay} baseline. We follow the convention and code released with VisPlay. Similar to Video-Zero, the method trains two co-evolving models, a Questioner and a Reasoner, both initialized from the same backbone (Qwen 3.5 9B), sequentially over three iterations. The Questioner first learns to write questions at the Reasoner's frontier. These questions are scored by the frozen Reasoner from the previous iteration (or the base model at initialization). The Questioner is then frozen to generate a new set of questions, which are pseudo-labelled by the same frozen Reasoner and used to train the next Reasoner. The training loop resembles $R_t \rightarrow Q_{t+1} \rightarrow D_{t+1} \rightarrow R_{t+1}$, where $t$ is the current iteration (starting from 0), $R_t$ the Reasoner, $Q_t$ the Questioner, and $D_t$ the questions generated by $Q_t$ to train the Reasoner. While the original work targets image reasoning, we adapt it to video clips and transcripts for social behaviour tasks. Gold labels are not used during training.

\begin{algorithm}[htbp]
\caption{VisPlay Co-Evolution}
\label{alg:visplay}
\begin{algorithmic}[1]
\Require Backbone $\theta_{0}$; train split $\mathcal{C}$; generation subset $\mathcal{C}_{\mathrm{gen}} \subset \mathcal{C}$ ($|\mathcal{C}_{\mathrm{gen}}| = 1{,}000$); validation split; iterations $T=3$
\Ensure Trained Reasoner $R$
\State $Q_{0}, R_{0} \gets \theta_{0}$
\For{$t = 0, \dots, T-1$}
    \State $Q_{t+1} \gets \textsc{GRPO}(Q_{t}, R_{t}, \mathcal{C}, r_{Q})$ \Comment{$R_{t}$ frozen; 20 optimizer steps} \label{alg:vp:q}
    \State $D_{t+1} \gets \textsc{GenerateQns}(Q_{t+1}, \mathcal{C}_{\mathrm{gen}})$ \Comment{$Q_{t+1}$ frozen; $N=8$ candidates per clip} \label{alg:vp:gen}
    \State $\widetilde{D}_{t+1} \gets \textsc{Filter}(\textsc{PseudoLabel}(R_{t}, D_{t+1}))$ \Comment{$R_{t}$ frozen; retain $0.3 \le c \le 0.8$} \label{alg:vp:filter}
    \State $R_{t+1} \gets \textsc{GRPO}(R_{t}, \widetilde{D}_{t+1}, r_{R})$ \Comment{20 optimizer steps} \label{alg:vp:r}
    \State Validate $R_{t+1}$; keep it if it improves on the best validation score
\EndFor
\State \Return best validation checkpoint of $R$
\end{algorithmic}
\end{algorithm}

\textbf{Training.} The key difference between VisPlay and Video-Zero is the number of iterations and update steps, as well as the emphasis on video grounding versus question diversity (see the Questioner reward below). We run one pipeline per task on the task's full training split and evaluate the checkpoint with the best validation score under the common evaluation procedure (Appendix~\ref{apx:evaluation_implementation}). The Questioner trains on the full split, capped at 20 optimizer steps per iteration (see Optimization and compute). Question generation for the Reasoner (line~\ref{alg:vp:gen}) reads a fixed, seeded subset of 1{,}000 clips from the same split to match the same number of images in VisPlay's released code. Each sample carries the unlabelled video clip and the transcript (wherever applicable). The PrefEval tasks have no video, and the transcript uses the user-assistant conversation rather than a single speaker's line.

The answer spaces are identical to Video-Zero's: a closed answer space constrains the Questioner's proposed answer and the pseudo-label vote to the task's labels (UR-FUNNY, MOSEI), while IntentQA and PrefEval are allowed to have open-ended free text for the pseudo-labelling stage ($\widetilde D_{t+1}$).

After each iteration (lines~\ref{alg:vp:q}--\ref{alg:vp:r}) we save the Reasoner checkpoint and score it on the full validation split of the task with the same settings as the test-split evaluation. The checkpoint replaces the current best if it strictly improves on the best validation score so far. Every run completes all three iterations to match the training pipeline of Visplay. The best validation checkpoint is the one evaluated on the test set of each task.


\begin{table}[tbp!]
    \centering
    \setlength{\belowcaptionskip}{6pt}
    \setlength{\tabcolsep}{4pt}
    \caption{VisPlay Reasoner checkpoints across iterations for \textbf{one} training run. Validation is the full validation split; test is the full test split. The tasks are reported in their corresponding metrics defined in Appendix~\ref{apx:evaluation_implementation}. \textbf{Bold} marks the iteration with the best validation score, which is the reported checkpoint; ties go to the earlier iteration. Only that checkpoint and the final one are test-evaluated. The results in Table~\ref{tab:main_results}, \ref{tab:results_with_std_dev} and \ref{tab:prefeval_full} uses the mean values across three runs.}
    \label{tab:visplay-ckpt}
    \small
    \begin{tabularx}{\textwidth}{@{}>{\raggedright\arraybackslash}X cc cc cc cc cc cc@{}}
        \toprule
        & \multicolumn{2}{c}{PrefEval EXP} & \multicolumn{2}{c}{PrefEval CHO} & \multicolumn{2}{c}{PrefEval PER} & \multicolumn{2}{c}{UR-FUNNY} & \multicolumn{2}{c}{MOSEI SEN} & \multicolumn{2}{c}{IntentQA} \\
        \cmidrule(lr){2-3} \cmidrule(lr){4-5} \cmidrule(lr){6-7} \cmidrule(lr){8-9} \cmidrule(lr){10-11} \cmidrule(lr){12-13}
        \textbf{Checkpoint} & Val & Test & Val & Test & Val & Test & Val & Test & Val & Test & Val & Test \\
        \midrule
        Iteration 1 & 38.00 & -- & \textbf{43.00} & \textbf{40.70} & \textbf{36.00} & \textbf{37.69} & 59.15 & -- & 62.70 & -- & 52.94 & -- \\
        Iteration 2 & \textbf{43.00} & \textbf{52.76} & 36.00 & -- & 36.00 & -- & \textbf{61.39} & \textbf{64.49} & \textbf{63.58} & \textbf{64.60} & 52.35 & -- \\
        Iteration 3 & 43.00 & 52.76 & 39.00 & 43.22 & 35.00 & 36.68 & 61.03 & 63.84 & 62.49 & 64.75 & \textbf{53.13} & \textbf{54.64} \\
        \bottomrule
    \end{tabularx}
\end{table}

\begin{figure}
    \centering
    \begin{promptbox}[breakable=false]{Questioner Template}
    The speaker said: '\{transcript\}'\\[4pt]
    You are an intelligent Question Generator. Your task is to create a question based on the given video clip and its transcript.\\[4pt]
    **Requirements (must follow exactly):**\\[4pt]
    1. Analyze the video clip and the transcript carefully and understand all details.\\
    2. Generate **exactly one question** that is directly related to the clip.\\
    3. Choose the question type from **only one** of the following:\\
    - \texttt{`multiple choice`} (only one correct answer, taken from the list of allowed answers below)\\
    4. The question must require analysis or reasoning, not just description.\\
    5. Provide the correct answer. The answer must be exactly one of: \{labels\}. Do NOT include any units or any extra text; provide only the option.\\
    6. **Output must be strictly in this format, with nothing else:**\\
    7. Question type must be **only one** of: \texttt{`multiple choice`}.\\[4pt]
    The following THREE blocks:\\
    <type>X</type>\\
    <question>Y</question>\\
    <answer>Z</answer>\\[4pt]
    **Strict rules:**\\
    - Do **not** use any other labels, punctuation, or formatting.\\
    - Do **not** add commentary, explanations, or extra text.\\
    - `X` must be exactly one of: \texttt{`multiple choice`}.\\
    - Always use the exact three-line structure above.\\
    - The answer must be exactly one of: \{labels\}. Do NOT include any units or any extra text; provide only the option.\\
    **Example of correct output:**\\
    <type>multiple choice</type>\\
    <question>Which of the allowed answers best describes the speaker in this clip?</question>\\
    <answer>\{label$_1$\}</answer>
\end{promptbox}
    \label{fig:prompt-questioner-visplay}
\end{figure}

\begin{figure}
    \centering
    \begin{promptbox}[breakable=false]{Reasoner Training Template}
        The speaker said: '\{transcript\}'\\[4pt]
        Please reason step by step carefully based on the question: \{question\} and the video clip. After completing your reasoning, you MUST output the final, clean, and concise answer strictly inside \textbackslash\textbackslash boxed\{ \}.The final answer MUST appear inside \textbackslash boxed\{\}, and nowhere else. If there is no boxed answer, your response is considered incorrect. You FIRST think about the reasoning process as an internal monologue and then provide the final answer. The reasoning process MUST BE enclosed within <think> </think> tags. The final answer MUST BE put in \textbackslash boxed\{\}.
    \end{promptbox}
    \label{fig:prompt-reasoner-visplay}
\end{figure}

\textbf{Visplay Questioner template.} The template adopts VisPlay's original prompt, replacing the image tag for the video clip and transcript.  The \{label]\} slot is specific to the task's answer space. In the original work, Visplay includes \texttt{numerical} and \texttt{regression} answer spaces. Since this work is concerned with social behavioral (hard to verify) tasks, we drop these answer spaces in the prompt. Instead, we include the \texttt{multiple choice} option and added an \texttt{open-ended} answer space for IntentQA. During generation (step 4 of Algorithm~\ref{alg:visplay}), the same template is followed by VisPlay's instruction \texttt{Generate one new, challenging reasoning question based on this video clip. Remember to format the output exactly as instructed.}

\textbf{Questioner reward.} Following VisPlay's released code, each rollout is scored
\begin{equation*}
    r_Q \;=\; r_{\mathrm{unc}} \;-\; r_{\mathrm{div}},
    \qquad
    r_{\mathrm{unc}} =
    \begin{cases}
    \min(c,\, 1-c) & \text{the three-block output parses,}\\[2pt]
    -1 & \text{otherwise.}
    \end{cases}
\end{equation*}

$c$ is the self-consistency of the frozen $R_t$ on the proposed question: $R_t$ answers it $m = 10$ times (temperature 1.0, top-$p$ 1.0, top-$k$ 40) through a scoring prompt with the answer \textbackslash boxed\{\} in. $c$ is computed by the fraction of the answers that is the majority answer. Every rollout counts in the denominator: a response with no boxed answer extracts a literal \texttt{None} placeholder and is scored accordingly. $r_{\mathrm{unc}}$ peaks at $c = 0.5$, rewarding questions that are the Reasoner's frontier.

$r_{\mathrm{div}}$ is VisPlay's repetition penalty. All questions in the reward batch are clustered by agglomerative clustering (average linkage, distance threshold 0.5) over pairwise distances $1 - \mathrm{BLEU}(q_i, q_j)$ (symmetrized sentence-BLEU with method-1 smoothing), and each rollout is penalized by the fraction of the batch in its own cluster. VisPlay computes this over reward batches of 4{,}096 questions; in our implementation, the reward batch is 240 (30 clips $\times$ 8 rollouts).

\textbf{Pseudo-labeling and filtering.} The frozen $Q_{t+1}$ generates $N = 8$ candidate questions per clip, and the frozen $R_t$ answers each candidate $m = 9$ times through VisPlay's prompt. The majority answer becomes the pseudo-label $\hat{y}$ with the self-consistency score computed as $c$. A candidate is retained iff $0.3 \le c \le 0.8$ (which is also similar to Video-Zero) and $\hat{y}$ is valid (e.g., not an empty string or None). VisPlay's implementation of hard drops also apply: on questions containing \texttt{box}, or with majority answers containing \texttt{text}).

\textbf{Reasoner's template.} We also adopt the template used by VisPlay. We excluded the method's thinking instructions that involve the <think> tags; the method and the evaluation in this work are run with thinking disabled in the backbone models. The text-only variants uses \texttt{transcript} instead of \texttt{video clip} in the template.

\textbf{Reasoner reward.} On the retained set (step 6 of Algorithm~\ref{alg:visplay}), each Reasoner rollout with boxed answer $\hat{a}$ is scored
\begin{equation*}
    r_R \;=\; 0.9 \cdot 1[\hat{a} = \hat{y}] \;+\; 0.1 \cdot 1[\mathrm{format}],
\end{equation*}
where the format term requires the whole response to be \textbackslash boxed\{...\}, and the accuracy term compares $\hat{a}$ against the pseudo-label $\hat{y}$ through the same parser of the other baselines (label parsing for closed spaces, normalized free text for IntentQA).

\textbf{Optimization and compute.} Both the Questioner and Reasoner train with GRPO on verl: group-normalized advantages, the clipped surrogate, and a KL loss against the rollout reference, with VisPlay's optimizer settings. Each role takes 20 optimizer steps per iteration with a rollout batch of 30 prompts and $G = 8$ rollouts each. The Questioner phase draws 600 clips of the training split to train itself. After which, the Questioner would generate questions from 1000 clips from the training set to train the reasoner; these 1000 questions will be filtered for formatting and for an appropriate difficulty level (neither too hard nor easy). The Reasoner is then trained on the retained (filtered) set of the generated questions, capped at 600 to maintain the compute budget. Over three iterations, that is 60 optimizer steps per role. VisPlay's own runs take 80 gradient updates per phase from their 20 rollout batches.

Rollouts sample at temperature 1.0 and top-$p$ 0.99. Thinking is off for both the Questioner and Solver. Video is sampled at 1 fps, 1 to 8 frames, resized to $176 \times 320$, to replicate the settings in the evaluation. Each run uses 3 H200 GPUs. During the Questioner phase one GPU hosts the frozen Reasoner as a vLLM scoring server and the other holds verl's rollout engine and trainer and the last card would seed the persistent rubrics. The four completed runs took roughly 12--33 hours each. Each iteration exports weights for both roles where only the Reasoner checkpoints are evaluated.

\begin{table}[htbp]
\centering
    \caption{Optimization settings for VisPlay. We follow the original work's released code, except the max response lengths, trimmed from 4096 to fit classification-length answers, and the video front end, which follows the evaluation configuration.}
    \label{tab:visplay-optimization-settings}
    \begin{tabular}{lcc}
    \toprule
    \textbf{Setting} & \textbf{Questioner} & \textbf{Reasoner} \\
    \midrule
    Optimization algorithm                        & GRPO               & GRPO               \\
    Learning rate                                 & $1 \times 10^{-6}$ & $1 \times 10^{-6}$ \\
    KL coefficient                                & 0.01               & 0.01               \\
    Iterations                                    & 3                  & 3                  \\
    Optimizer steps per iteration                 & 20                 & 20                 \\
    Prompts per rollout batch                     & 30                 & 30                 \\
    Rollouts per prompt for GRPO                  & 8                  & 8                  \\
    Optimizer steps per rollout batch             & 1                  & 1                  \\
    Rollout temperature / top-$p$                 & 1.0 / 0.99         & 1.0 / 0.99         \\
    Frozen-Reasoner rollouts for scoring ($m$)    & 10                 & --                 \\
    Generation clips per iteration                & 1{,}000            & --                 \\
    Candidate questions per clip ($N$)            & 8                  & --                 \\
    Rollouts for pseudo-labeling ($m$)            & --                 & 9                  \\
    Retention gate on confidence $c$              & --                 & $[0.3,\, 0.8]$     \\
    Checkpoint used for test                      & --                 & best val iteration \\
    Max response length                           & 2048               & 1024               \\
    Training precision                            & bfloat16           & bfloat16           \\
    Rollout engine                                & vLLM               & vLLM               \\
    \bottomrule
    \end{tabular}
\end{table}

\textbf{Ensuring a fair implementation for VisPlay.} Our implementation runs the authors' released training script and keeps every setting that does not depend on the corpus: learning rate, KL coefficient, $G = 8$, sampling temperature and top-$p$, the difficulty retention window, the format-reward weight, and the clustering settings of the diversity term (Table~\ref{tab:visplay-optimization-settings}), together with the three-iteration schedule of 20 optimizer steps per role. The one corpus-dependent setting is the rollout batch. VisPlay's released configuration draws 512 prompts per step from a corpus of several thousand \textbf{images}; our training splits are smaller (and may contain video clips sampled into multiple frames). To address this, we draw 30 prompts per step so that the 20 steps of a phase cover its 600 clips exactly once. The number of gradient updates per phase therefore matches the original work and only the number of prompts consumed per update follows the corpus. Where we deviate from the original work, we attempt to do so in VisPlay's favour: the Questioner draws its clips from the whole training split at every iteration rather than from a fixed subset, and the Reasoner is validated on the full validation split after every iteration with the best checkpoint evaluated on the test split. Note that the original work evaluates only the last iteration. Table~\ref{tab:visplay-ckpt} reports both. In compute, a VisPlay run takes 26-44 GPU hours, three to six times the GPU-hours of RLCE.

\clearpage

\subsection{Video-Zero}

This section details how we implement the Video-Zero baseline. The method trains two co-evolving models, a Questioner and a Solver, both initialized from the same backbone (Qwen 3.5 9B), sequentially over up to five iterations. The Questioner first learns to write questions about temporally localized evidence in unlabelled video clips, scored by the frozen Solver from the previous iteration (or the base model at initialization). For tasks without video clips (e.g., PrefEval), the Questioner instead writes questions about the transcript alone, and the parts of the method that depend on video are switched off. After training, the Questioner is then frozen to generate a new set of questions, which are pseudo-labelled by the same frozen Solver and used to train the next Solver. The training loop resembles $S_t \rightarrow Q_{t+1} \rightarrow D_{t+1} \rightarrow S_{t+1}$, where $t$ is the current iteration (starting from 0), $S_t$ the Solver, $Q_t$ the Questioner, and $D_t$ the questions generated by $Q_t$ to train the Solver. Gold labels are not used during training.

\begin{algorithm}[tbp]
\caption{Video-Zero Co-Evolution}
\label{alg:video-zero}
\begin{algorithmic}[1]
\Require Backbone $\theta_{0}$; clip subset $\mathcal{C}$ ($|\mathcal{C}| = 600$); validation split; iterations $T \le 5$
\Ensure Trained Solver $S$
\State $Q_{0}, S_{0} \gets \theta_{0}$
\For{$t = 0, \dots, T-1$}
    \State $Q_{t+1} \gets \textsc{GRPO}(Q_{t}, S_{t}, \mathcal{C}, r_{Q})$ \Comment{$S_{t}$ frozen; 2 optimizer steps} \label{alg:vz:q}
    \State $D_{t+1} \gets \textsc{GenerateQns}(Q_{t+1}, \mathcal{C})$ \Comment{$Q_{t+1}$ frozen; one question per clip} \label{alg:vz:gen}
    \State $\widetilde{D}_{t+1} \gets \textsc{Filter}(\textsc{PseudoLabel}(S_{t}, D_{t+1}))$ \Comment{$S_{t}$ frozen; retain by confidence gate} \label{alg:vz:filter}
    \State $S_{t+1} \gets \textsc{GRPO}(S_{t}, \widetilde{D}_{t+1}, r_{S})$ \Comment{one optimizer step per 256 retained items} \label{alg:vz:s}
    \State Validate $S_{t+1}$ with early stopping (no improvements over 2 iterations.)
\EndFor
\State \Return best validation checkpoint of $S$
\end{algorithmic}
\end{algorithm}

\textbf{Training.} We run one pipeline per task. Each run draws 600 clips at random from the task's training split (seed 0) and reuses the same 600 clips in every iteration following the paper's training recipe; This is standardized for both Questioner training (line~\ref{alg:vz:q}) and for generating questions for the solver (line~\ref{alg:vz:gen}). Each sample carries the unlabelled video clip and the transcript (wherever applicable). The PrefEval tasks have no video, so the pipeline runs on the transcript alone: the Questioner template asks for a question about the transcript and outputs only the question and answer tags, the Solver template says \texttt{Read the transcript} and drops the grounding tag, and the evidence term of the Questioner reward is switched off (Figure~\ref{fig:prompt-textonly-vidzero}). The PrefEval transcript is the user-assistant conversation rather than a single speaker's line.

The answer space is set per task: a closed answer space constrains the Questioner's proposed answer and the pseudo-label vote to the task's labels (UR-FUNNY, MOSEI), while IntentQA and PrefEval use open-ended free text.

After each iteration (lines~\ref{alg:vz:q}--\ref{alg:vz:s}) we save the Solver checkpoint and score it on the full validation split of the task with the same settings as the test-split evaluation. The checkpoint replaces the current best if it strictly improves on the best validation score so far. A run stops early when the validation score has not improved for two iterations in a row after the first, and otherwise runs the full five iterations. The best validation checkpoint is selected for evaluation on the test set of each task. 

\begin{table}[htbp!]
    \centering
    \setlength{\belowcaptionskip}{6pt}
    \setlength{\tabcolsep}{4pt}
    \caption{Video-Zero Solver checkpoints across iterations for \textbf{one} full training run. Validation is the full validation split; test is the full test split. The tasks are reported in their corresponding metrics defined in Appendix~\ref{apx:evaluation_implementation}. \textbf{Bold} marks the iteration with the best validation score, which is the reported checkpoint; ties go to the earlier iteration. Only that checkpoint, the final one, and iteration 4 (the fixed iteration count of the original work) are test-evaluated. Rows after an early stop (patience 2) are left empty. The results in Table~\ref{tab:main_results}, \ref{tab:results_with_std_dev} and \ref{tab:prefeval_full} uses the mean values across three runs}
    \label{tab:video-zero-ckpt}
    \small
    \begin{tabularx}{\textwidth}{@{}>{\raggedright\arraybackslash}X cc cc cc cc cc cc@{}}
        \toprule
        & \multicolumn{2}{c}{PrefEval EXP} & \multicolumn{2}{c}{PrefEval CHO} & \multicolumn{2}{c}{PrefEval PER} & \multicolumn{2}{c}{UR-FUNNY} & \multicolumn{2}{c}{MOSEI SEN} & \multicolumn{2}{c}{IntentQA} \\
        \cmidrule(lr){2-3} \cmidrule(lr){4-5} \cmidrule(lr){6-7} \cmidrule(lr){8-9} \cmidrule(lr){10-11} \cmidrule(lr){12-13}
        \textbf{Checkpoint} & Val & Test & Val & Test & Val & Test & Val & Test & Val & Test & Val & Test \\
        \midrule
        Iteration 1 & 40.00 & -- & 43.00 & -- & 37.00 & -- & \textbf{56.39} & \textbf{62.26} & 63.01 & -- & 52.69 & -- \\
        Iteration 2 & 42.00 & -- & \textbf{45.00} & \textbf{39.70} & 38.00 & -- & 54.73 & -- & 63.17 & -- & 52.30 & -- \\
        Iteration 3 & \textbf{43.00} & \textbf{53.27} & 43.00 & -- & 37.00 & -- & 54.65 & -- & \textbf{63.29} & \textbf{64.65} & 52.40 & -- \\
        Iteration 4 & 41.00 & 53.27 & 44.00 & 44.22 & \textbf{39.00} & \textbf{38.69} & 53.92 & 62.79 & 63.16 & 64.93 & 52.98 & 54.83 \\
        Iteration 5 & 43.00 & 49.75 & -- & -- & 39.00 & 36.68 & -- & -- & 63.19 & 64.97 & \textbf{53.13} & \textbf{54.97} \\
        \bottomrule
    \end{tabularx}
\end{table}

\begin{figure}
    \centering
    \begin{promptbox}[breakable=false]{Questioner Template}
    The speaker said: '\{transcript\}'\\[4pt]
    You are given an unlabeled video. Your job is to create one challenging question about \{subject\} that can only be answered by watching a specific moment in the video, not from general knowledge or a single static frame.\\[4pt]
    Work in this order:\\
    1. Find the most informative temporal evidence span (a start and end time in seconds).\\
    2. Describe what happens in that span.\\
    3. Write a question about \{subject\} during that span, answerable with exactly one of: \{labels\}.\\
    4. Give the correct answer to your own question. The answer must be exactly one of: \{labels\}.\\[4pt]
    Output exactly:\\
    <evidence>[start, end]</evidence>\\
    <description>...</description>\\
    <question>...</question>\\
    <answer>...</answer>
\end{promptbox}
    \label{fig:prompt-questioner-vidzero}
\end{figure}

\begin{figure}
    \centering
    \begin{promptbox}[breakable=false]{Text-Only Questioner Template}
    The speaker said: '\{transcript\}'\\[4pt]
    You are given an unlabeled transcript of a speaker. Your job is to create one challenging question about \{subject\} that can only be answered by reading what the speaker actually said, not from general knowledge.\\[4pt]
    Work in this order:\\
    1. Write a question about \{subject\}, answerable with exactly one of: \{labels\}.\\
    2. Give the correct answer to your own question. The answer must be exactly one of: \{labels\}.\\[4pt]
    Output exactly:\\
    <question>...</question>\\
    <answer>...</answer>
    \end{promptbox}
    \begin{promptbox}[breakable=false]{Text-Only Solver Template}
    The speaker said: '\{transcript\}'\\[4pt]
    Read the transcript and answer the following question.\\[4pt]
    Question: \{question\}\\[4pt]
    Give your final answer inside <answer> </answer> tags. Answer with exactly one of: \{labels\}.\\[4pt]
    Your output format must be:\\
    <answer>...</answer>
    \end{promptbox}
    \label{fig:prompt-textonly-vidzero}
\end{figure}

\textbf{Video-Zero Questioner template.} We adapt Video-Zero's four-step prompt in our implementation. The \{subject\} slot is \texttt{the speaker's sentiment} for MOSEI and \texttt{whether the speaker is being humorous} for UR-FUNNY. The \{labels\} slot is the task's answer space. The open-ended template (IntentQA) differs in three places: the first sentence says \texttt{about it} instead of \texttt{about \{subject\}}, step 3 reads \texttt{Write a question whose answer depends on that span.}, and step 4 drops the label sentence. The text-only template for PrefEval uses \texttt{the speaker} as the subject and drops both label sentences. During generation (line~\ref{alg:vz:gen}) the same template is used and the Questioner writes one question per clip.

\textbf{Questioner reward.} Following Video-Zero's appendix, each rollout is scored:
\begin{equation*}
r_Q \;=\; 0.1\, r_{\mathrm{format}} \;+\; 0.5\, r_{\mathrm{learn}} \;+\; 0.3\, r_{\mathrm{dep}} \;+\; \min(r_{\mathrm{evid}},\, 0.3),
\end{equation*}

$r_{\mathrm{format}}$ is 1 when all four output tags parse and 0 otherwise; in the closed-space runs the proposed answer must additionally parse to exactly one label of the space.

$r_{\mathrm{learn}}$: the frozen $S_t$ answers the proposed question $m = 5$ times with the clip, and $c$ is the fraction of those answers that agree with their own majority answer. Open-ended, $r_{\mathrm{learn}} = \min(c, 1 - c)$, peaking at $c = 0.5$. Under a closed $K$-label space $c$ cannot fall below $1/K$, so we replace it with a triangle over $[1/K, 1]$: zero at both ends, rising linearly to a peak of $0.5$ at the midpoint. We set $m$ = 5 to match the order of its other rollout counts.

$r_{\mathrm{dep}} = \max(0,\, p_{\mathrm{with}} - p_{\mathrm{without}})$, where $p_{\mathrm{with}}$ is the fraction of the 5 with-clip answers that match the Questioner's own answer and $p_{\mathrm{without}}$ the same over 5 queries without the video clips. The blind query removes both the frames and the transcript and prepends \texttt{No video is available for this question. Answer as best you can from the question text alone.}, so $r_{\mathrm{dep}}$ measures dependency on the whole clip and the transcript against a question-only baseline. For the text-only tasks, we use only the no video clip prefix instead.

$r_{\mathrm{evid}} = s_{\mathrm{span}} + s_{\mathrm{event}} + s_{\mathrm{temp}}$. The original work did not have sufficient information on these term. As such, we implement these terms in the following manner: 
\begin{itemize}
    \item $s_{\mathrm{span}} = 1 - \mathrm{coverage}/0.9$ for a well-formed span inside $[0, T]$ ($T$ the clip length, coverage its fraction of $T$), and 0 for an invalid or out-of-range span or one covering at least 90\% of the clip, which suppresses degenerate whole-clip spans.
    \item $s_{\mathrm{event}}$ and $s_{\mathrm{temp}}$ count how many entries of a small lexicon of action cues (31 entries) and temporal cues (19 entries) appear in the lower-cased description, each scored $\min(\mathrm{hits}, 2)/2$. The exact lexicons will be provided in the GitHub repository for this work.
\end{itemize}

\textbf{Pseudo-labelling and filtering.} For each generated question the frozen $S_t$ answers $m = 10$ times through the Solver template. The majority answer becomes the pseudo-label $\hat{y}$ and $c$ is its share of the 10 rollouts. In the open-ended runs a question is retained iff $0.3 \le c \le 0.8$, as in the original work. In the closed-space runs a chance-level majority is easy to reach with few labels, so the lower bound is raised to $\min(0.6,\, 1/K + 0.2)$ for a $K$-label space, the upper bound stays at $0.8$, and a question whose top answer leads the runner-up by fewer than two votes is rejected. The target span $\hat{\tau}$ is the median start and median end over the rollouts that agree with $\hat{y}$; if none of those spans is usable, the Questioner's own evidence span is used instead as detailed in Video Zero's training recipe.

\textbf{Solver template.} We adopt the Solver prompt of Video-Zero. The sentence \texttt{Answer with exactly one of: \{labels\}.} appears only in the closed-space runs; the open-ended runs omit it. Thinking is disabled in both roles, so the templates carry no <think> instruction.

\begin{figure}
    \centering
    \begin{promptbox}[breakable=false]{Solver Training Template}
        The speaker said: '\{transcript\}'\\[4pt]
        Watch the video and answer the following question.\\[4pt]
        Question: \{question\}\\[4pt]
        Give your final answer inside <answer> </answer> tags. Answer with exactly one of: \{labels\}. Then, on a new line, give the time span in seconds that your answer relies on inside <grounding> </grounding> tags.\\[4pt]
        Your output format must be:\\
        <answer>...</answer>\\
        <grounding>[start, end]</grounding>
    \end{promptbox}
    \label{fig:prompt-solver-vidzero}
\end{figure}

\textbf{Solver reward.} On the retained set (step 5 of Algorithm~\ref{alg:video-zero}), each Solver rollout with answer $\hat{a}$ and predicted span $\tau_p$ is scored
\begin{equation*}
r_S = 1[\hat{a} = \hat{y}] \,\bigl(1 + 0.5\, r_{\mathrm{align}}\bigr),
\qquad
r_{\mathrm{align}} = \mathrm{IoU}(\tau_p, \hat{\tau}) \prod_{b \in \{\mathrm{start}, \mathrm{end}\}} \max\!\Bigl(0,\; 1 - \frac{|\tau_{p,b} - \hat{\tau}_b|}{T}\Bigr).
\end{equation*}
An incorrect answer scores 0 regardless of the span; a correct answer earns the base reward, multiplied up when its span aligns with the consensus target. Spans are parsed leniently (brackets optional; ``12 to 18 seconds'' also parses) but scored strictly: a predicted or target span outside $[0, T]$, or with end $\le$ start, earns zero alignment reward.

\textbf{Optimization and compute.} Both roles train with GRPO on verl: group-normalized advantages, the clipped surrogate, and a KL loss against the rollout reference. Per iteration the Questioner takes 2 optimizer steps (600 prompts, mini-batch 300) and the Solver one (the full retained set). 

Over five iterations that is 10 and 5 optimizer steps. Rollouts run in vLLM at temperature 1.0 in bfloat16. Frames are sampled at 2 fps up to 32 frames under a total budget of $2048 \cdot 28 \cdot 28$ pixels per clip, which keeps the video prompt near the paper's 1790 tokens on Qwen3.5's coarser vision-token grid regardless of clip length. Each iteration exports weights for both roles and we only evaluate the selected Solver checkpoint with the highest validation score for each task.


\begin{table}[tbp]
\centering
    \caption{We used the same optimization settings in Video-Zero.}
    \label{tab:video-zero-optimization-settings}
    \begin{tabular}{lcc}
    \toprule
    \textbf{Setting} & \textbf{Questioner} & \textbf{Solver} \\
    \midrule
    Optimization algorithm       & GRPO               & GRPO               \\
    Learning rate                & $1 \times 10^{-6}$ & $1 \times 10^{-6}$ \\
    Epochs per iteration         & 1                  & 1                  \\
    KL coefficient               & 0.01               & 0.01               \\
    Rollouts per prompt for GRPO & 4                  & 5                  \\
    Rollouts for pseudo-labeling & --                 & 10                 \\
    Max response length          & 4096               & 2048               \\
    Training precision           & bfloat16           & bfloat16           \\
    Rollout engine               & vLLM               & vLLM               \\
    \bottomrule
    \end{tabular}
\end{table}

\textbf{Ensuring a fair implementation for Video-Zero.} In Video-Zero, the authors released the inference and evaluation code. We reconstruct training from their paper and re-use their training parameters. The reconstruction follows the paper's Eqs.~1--6 and Table~6 term by term. The two terms the paper does not fully specify ($s_{\mathrm{event}}$ and $s_{\mathrm{temp}}$) are documented above, and the full training code will be released with ours so that our reconstruction can be verified. We also re-use the paper's five-iteration schedule and its corpus size of 600 clips, drawn from the training split and fixed across the training iterations. The paper's own scaling ablation claims that 600, 1000, and 2000 clips produce near-identical results. Where we deviate from the original work, we do so in Video-Zero's favour. We give the transcripts to both roles (the Questioner and Solver), since several of our tasks depend on what is said. On the closed-label tasks we let the Solver answer in the label space and make the learnability frontier $K$-aware. That is, we adjust the frontier based on $K$ classes. The motivation for this change was to adjust the difficult according to the number of classes for pseudo-labelling--- with $K = 3$ the paper's fixed window would pass almost every noisy question. In addition, the paper reports iteration 4 only; we validate the Solver on the full validation split after every iteration, then evaluate the best checkpoint on the test split. We report the selected checkpoint, the final one, and iteration 4 (the paper's own comparison point) in Table~\ref{tab:video-zero-ckpt}. This means that Video-Zero is evaluated at its best validation checkpoint. In compute, a Video-Zero run takes 28--40 GPU hours, three to five times the GPU-hours of \method. These efforts ensure that Video-Zero is a non-trivial baseline for \method to compare against.


\subsection{RLAC}
\label{apx:rlac_implementation}

This method uses the same backbone model (e.g., Qwen 3.5 9B) for three roles: a
Generator to be trained; a Critic, also trained, that attacks the Generator's
answers; and a frozen Validator that adjudicates between them. Unlike the other
three baselines, both players are optimized. Their rewards are inverted, so the
Critic is rewarded when the Generator is penalized, and the verifier adapts to
the policy instead of staying fixed. The original method targets
free-form factual generation and code, where verification is grounded in
Wikipedia through FactScore or in executing a reference test case. We adapt it
to social understanding tasks, where neither is available.

Each round has two GRPO phases over the same clip corpus. In the Generator
phase, the Critic and Validator are frozen and every Generator rollout is
attacked. The Critic reads the clip, the question and the response, and returns
the single claim the response most likely gets wrong. In the Critic phase the
Generator is frozen and the Critic trains on pairs harvested from the previous
phase. The loop is $C_t \rightarrow G_{t+1} \rightarrow P_{t+1} \rightarrow
C_{t+1}$, where $t$ is the round, $G_t$ the Generator, $C_t$ the Critic, and
$P_t$ the pairs harvested to train it. We do not include the gold label anywhere
in this pipeline.

\begin{algorithm}[htbp]
\caption{RLAC Adversarial Co-Evolution}
\label{alg:rlac}
\begin{algorithmic}[1]
\Require Backbone $\theta_{0}$; clip corpus $\mathcal{C}$; rounds $T=5$
\Ensure Final Generator $G_{T}$
\State $G_{0}, C_{0} \gets \theta_{0}$; \; Validator $V \gets \theta_{0}$ \Comment{$V$ frozen throughout}
\For{$t = 0, \dots, T-1$}
    \State $G_{t+1} \gets \textsc{GRPO}(G_{t}, \mathcal{C}, r_{G})$ \Comment{$C_{t}, V$ frozen; $r_G = 1 - R$}
    \State $P_{t+1} \gets \textsc{Harvest}(G_{t+1}, \mathcal{C})[:M]$ \Comment{one pair per clip, see \textbf{Pair cap}}
    \State $C_{t+1} \gets \textsc{GRPO}(C_{t}, P_{t+1}, r_{C})$ \Comment{$G_{t+1}, V$ frozen; $r_C = R$}
\EndFor
\State \Return $G_{T}$
\end{algorithmic}
\end{algorithm}

\textbf{Training.} We perform one RLAC run per task on the same training
corpora as the other baselines and evaluate it with the same protocol
(Appendix~\ref{apx:evaluation_implementation}).

The \{evidence\} slot is \texttt{the video clip and the transcript of what is
said}, or \texttt{the video clip} alone where the dump carries no transcript.
The original method numbers the response's sentences and asks the critic for an
index. We ask for the quoted \emph{span} instead, because our responses are a
label plus one to three sentences of justification, and the error is usually a
phrase inside a sentence and not a whole sentence. Quoting the span also lets
the meta-verification below check for an exact substring before falling back
to a paraphrase check.

\textbf{Meta-verification.} Before any claim reaches the Validator, a frozen
meta-verifier checks that the claim actually originates in the quoted span.
Without this step the Critic can be rewarded for claims that the response never
made, which the original work calls exploitation through information injection.
This role is text-only. It compares two strings and never sees the clip.

\begin{figure}
    \centering
    \begin{promptbox}[breakable=false]{Critic Template}
        You are a strict critic. You are given \{evidence\}, one question about it, and one candidate response.\\[4pt]
        Find the single claim in the response that is most clearly NOT supported by \{evidence\_ref\}. A claim is anything the response asserts or takes for granted, including what it assumes about the speaker or the user. Pick the one you are most confident is wrong; do not list several, and do not object to the response's tone, length or style.\\[4pt]
        \#\# Question\\
        \{question\}\\[4pt]
        \#\# Transcript\\
        \{transcript\}\\[4pt]
        \#\# Candidate response (one of: \{labels\})\\
        \{response\}\\[4pt]
        Reply in exactly three lines and nothing else:\\
        reason: <one sentence saying what is wrong>\\
        span: <the exact wording copied from the response that carries it>\\
        claim: <a standalone statement of no more than 20 words, stating what the response claims -- not what is true>
    \end{promptbox}
    \label{fig:prompt-critic-rlac}
\end{figure}
\begin{figure}
    \centering
    \begin{promptbox}[breakable=false]{Meta-Verifier Template}
        You are an assistant that checks whether an extracted claim truly originates from a given piece of text. Only decide whether it is verbatim, or a direct paraphrase of part of it. Do not judge whether the claim is true.\\[4pt]
        Text: ``\{span\}''\\
        Extracted claim: ``\{claim\}''\\[4pt]
        Reply in exactly two lines and nothing else:\\
        reason: <one sentence>\\
        verdict: yes / no
    \end{promptbox}
    \label{fig:prompt-meta-rlac}
\end{figure}

\textbf{Validation.} The surviving claim goes to the frozen Validator with the
clip and the transcript. It does \emph{not} see the response the claim was
extracted from, and it is not told that a Critic proposed it, since knowing that
an adversary selected the claim would bias it toward ``not supported''. The
Validator is the component that departs most from the original. The original
validates against Wikipedia or by executing a test. A label-free corpus of
social behaviour offers neither, so we use the frozen backbone as the Validator
instead.

\begin{figure}
    \centering
    \begin{promptbox}[breakable=false]{Validator Template}
        You are checking one statement against \{evidence\}.\\[4pt]
        \#\# Question under discussion\\
        \{question\}\\[4pt]
        \#\# Transcript\\
        \{transcript\}\\[4pt]
        \#\# Statement\\
        \{claim\}\\[4pt]
        Using only \{evidence\_ref\}, is the statement supported? Answer yes when it is clearly supported, and no when it is contradicted or the evidence does not support it.\\[4pt]
        Reply in exactly two lines and nothing else:\\
        reason: <one sentence>\\
        verdict: yes / no
    \end{promptbox}
    \label{fig:prompt-validator-rlac}
\end{figure}

\textbf{Rewards.} Both players are scored from the same quantity. Writing $u$ for the event
that a proposal survives meta-verification and is then ruled \emph{unsupported},
each exchange is scored
\begin{equation*}
    R \;=\; 0.8 \cdot 1[u] \;+\; 0.2 \cdot 1[\mathrm{format}],
    \qquad r_C \;=\; R, \qquad r_G \;=\; 1 - R ,
\end{equation*}
where the format term checks the reply shape of each role: the
\texttt{<answer>}/\texttt{<justification>} pair for the Generator and the three
labelled lines for the Critic. The correctness and format terms are summed and
not gated, so a well-formed rollout that loses the exchange still scores higher
than a malformed one in the same GRPO group. When a proposal reaches no verdict
(the Critic emits nothing parseable, or a verifier returns no decision), the
Critic scores $0$, and the Generator receives its group's mean so that its
advantage is zero and it is not penalized for the verifier's failure. The
original updates both players with DPO and states that the method is agnostic
to the choice of policy optimization algorithm, naming GRPO as an option.

\textbf{Pair cap.} In the original method, the Critic's dataset is built over
\emph{every} rollout of every clip. At 600 clips and $G=8$ Generator rollouts
this is 4{,}608 pairs per round with no deduplication. Under GRPO that is 144
Critic optimizer steps per round against the Generator's 18, so the Critic
phase alone would cost roughly four times the Generator. We cap the harvest at $M=600$, one pair per
clip, which gives the two players 18 steps each and brings a run to about 30
H200 GPU-hours. This is a deviation from the original method to ensure the compute
budget is comparable across methods. Its cost is that the Critic trains on a
sample of the Generator's outputs instead of the full set.

\textbf{Optimization and compute.} Both roles train with GRPO on verl:
group-normalized advantages, the clipped surrogate, and a KL loss against the
rollout reference. The Generator takes 18 optimizer steps per round (600 prompts,
train batch 32, $G=8$ rollouts) and the Critic 18 (600 pairs, train batch 32,
$G=4$); over five rounds that is 90 optimizer steps per role. Rollouts run in
vLLM at temperature 1.0 in bfloat16. Frames are sampled at 2 fps up to 32 frames
under a total budget of $2048 \cdot 28 \cdot 28$ pixels per clip, manifestly more than our method \method (8 frames). Table~\ref{tab:rlac-optimization-settings}
lists the full settings for both players.

\begin{table}[htbp]
\centering
    \caption{Optimization settings for RLAC. The original work trains both
    players with DPO but states that the method is agnostic to the policy optimization algorithm, e.g., GRPO. We use GRPO with the same batch size, group size, learning rate and KL coefficient as the other
    baselines. The reward includes a format term, which DPO does not need because an
    unparseable response is left unpaired. Generator and Critic are trained in
    alternating phases within each round from the same Qwen 3.5 9B backbone,
    which also serves as the frozen Validator.}
    \label{tab:rlac-optimization-settings}
    \begin{tabular}{lcc}
    \toprule
    \textbf{Setting} & \textbf{Generator} & \textbf{Critic} \\
    \midrule
    Optimization algorithm                        & GRPO               & GRPO               \\
    Learning rate                                 & $1 \times 10^{-6}$ & $1 \times 10^{-6}$ \\
    KL coefficient                                & 0.01               & 0.01               \\
    Adversarial rounds                            & 5                  & 5                  \\
    Optimizer steps per round                     & 18                 & 18                 \\
    Prompts per rollout batch                     & 32 clips           & 32 pairs           \\
    Rollouts per prompt for GRPO                  & 8 ($K$)            & 8 ($N$)            \\
    Optimizer steps per rollout batch             & 1                  & 1                  \\
    Rollout temperature / top-$p$                 & 1.0 / 0.99          & 1.0 / 0.99          \\
    Critic-server proposal temperature            & 1.2                & --                 \\
    Frozen-validator temperature / max tokens     & 0.0 / 256           & 0.0 / 256           \\
    Max prompt length                             & 4096         & 6144        \\
    Max response length                           & 2048         & 1024                \\
     Checkpoint used for test                      & best val iteration        & --                 \\
    Training precision                            & bfloat16           & bfloat16           \\
    Rollout engine                                & vLLM               & vLLM               \\
    \bottomrule
    \end{tabular}
\end{table}

\textbf{Ensuring a fair implementation for RLAC.} The original work has no
official code release, but the code attached to its OpenReview submission is
available and we build on it. We keep its structure: the three roles from one backbone, the Critic that names
one unsupported claim per response, the meta-verifier that blocks claims the
response never made, and the frozen Validator that never sees the response and
is not told that a Critic proposed the claim. We keep the five-round
alternating schedule and use the same 600-clip corpus per round as Video-Zero.
The original trains both players with DPO and does not report batch sizes, so
we use GRPO, which the original work names as an admissible choice, with the same
learning rate, KL coefficient, batch size and group size as the other baselines
(Table~\ref{tab:rlac-optimization-settings}). Where we depart from the
original, we attempt to do so in RLAC's favour or to keep the compute comparable. The
Validator receives the clip and the transcript, the same evidence as the
Generator. The reward includes a format term so that a malformed rollout does
not collapse its GRPO group, and when a verifier returns no verdict the
Generator receives its group's mean so that it is not penalized for the
verifier's failure. The Critic is trained on one pair per clip so that both
players take the same number of optimizer steps. Over five rounds each player takes 90 optimizer steps. In compute, an RLAC run takes about 30 H200 GPU-hours. These efforts ensure that RLAC is a non-trivial baseline for \method to compare against.



\end{document}

\end{document}